\documentclass[10pt,twocolumn,letterpaper]{article}

\usepackage{iccv}
\makeatletter
\@namedef
{ver@everyshi.sty}{}
\makeatother
\usepackage{cancel,bbm}
\usepackage{tikz}
\usepackage{times}
\usepackage{epsfig}
\usepackage{graphicx,tipa}
\usepackage{amsmath}
\usepackage{amssymb}
\usepackage{amsthm}
\usepackage{caption}
\usepackage{subcaption}
\usepackage{booktabs}
\usepackage{svg,svg-extract}
\usepackage[linesnumbered,ruled,vlined]{algorithm2e}
\usepackage{comment}
\newtheorem{innercustomthm}{Definition}
\newenvironment{customthm}[1]
  {\renewcommand\theinnercustomthm{#1}\innercustomthm}
  {\endinnercustomthm}

\newtheorem{innercustomprop}{Proposition}
\newenvironment{customprop}[1]
  {\renewcommand\theinnercustomprop{#1}\innercustomprop}
  {\endinnercustomprop}

\DeclareFontFamily{OMX}{yhex}{}
\DeclareFontShape{OMX}{yhex}{m}{n}{<->yhcmex10}{}
\DeclareSymbolFont{yhlargesymbols}{OMX}{yhex}{m}{n}
\DeclareMathAccent{\arc}{\mathord}{yhlargesymbols}{"F3}

\usepackage[pagebackref=true,breaklinks=true,letterpaper=true,colorlinks,bookmarks=false]{hyperref}

 \iccvfinalcopy 

\def\iccvPaperID{9285} 

\ificcvfinal\fi

\begin{document}

\title{LoOp: Looking for Optimal Hard Negative\\ Embeddings for Deep Metric Learning}

\author{Bhavya Vasudeva$^1$\thanks{Equal contribution} \quad Puneesh Deora$^1$\footnotemark[1] \quad Saumik Bhattacharya$^2$ \quad Umapada Pal$^1$ \quad Sukalpa Chanda$^3$\\
$^1$Indian Statistical Institute, Kolkata, India \quad
$^2$Indian Institute of Technology, Kharagpur, India\\
$^3$\O{}stfold University College, Halden, Norway}

\maketitle
\ificcvfinal\thispagestyle{empty}\fi

\begin{abstract}
Deep metric learning has been effectively used to learn distance metrics for different visual tasks like image retrieval, clustering, etc. In order to aid the training process, existing methods either use a hard mining strategy to extract the most informative samples or seek to generate hard synthetics using an additional network. Such approaches face different challenges and can lead to biased embeddings in the former case, and (i) harder optimization (ii) slower training speed (iii) higher model complexity in the latter case. In order to overcome these challenges, we propose a novel approach that looks for optimal hard negatives (LoOp) in the embedding space, taking full advantage of each tuple by calculating the minimum distance between a pair of positives and a pair of negatives. Unlike mining-based methods, our approach considers the entire space between pairs of embeddings to calculate the optimal hard negatives. Extensive experiments combining our approach and representative metric learning losses reveal a significant boost in performance on three benchmark datasets\footnote{Code available at \url{https://github.com/puneesh00/LoOp}}.
\end{abstract}
\vspace{-2mm}
\section{Introduction}
\vspace{-1mm}
Deep metric learning tries to learn an embedding space, where closeness between embeddings encodes the level of semantic similarities between the data samples. This is done by leveraging deep neural networks to learn the mapping between the data samples and the embedding space and enforcing the embeddings belonging to the same class to lie close while pushing those belonging to different classes further apart. Deep metric learning-based methods have achieved state-of-the-art (SOTA) results for several tasks, like face recognition \cite{cont1,triplet1,fr1,fr2}, re-identification \cite{rid1,rid2,rid3,rid4}, image retrieval \cite{ret2,ret3}, etc.
\begin{figure}[h!]
    \centering
    \includegraphics[scale=0.35]{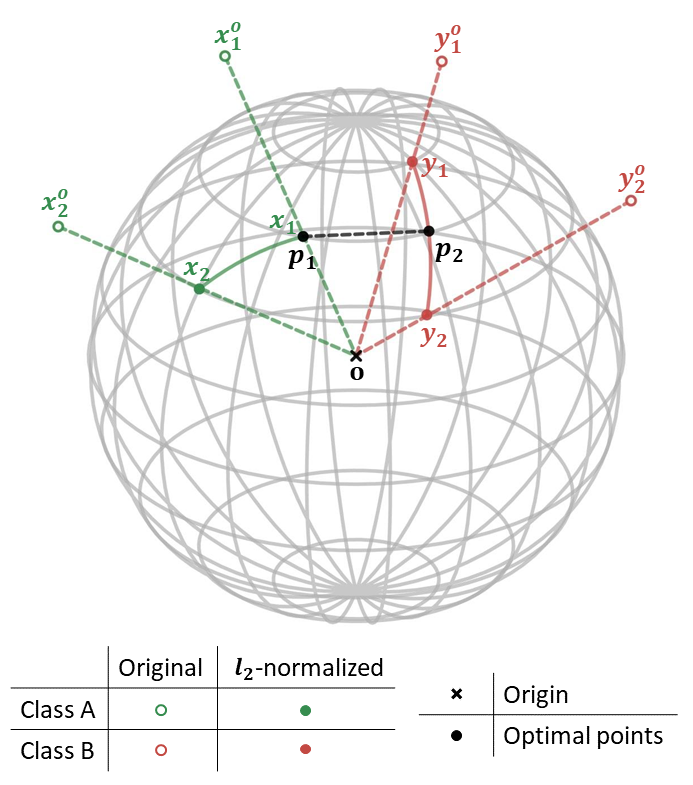}
    \caption{Illustration of the proposed approach. Given two pairs of points in the embedding space, $\mathbf{x_1}$, $\mathbf{x_2}$ from class A, and $\mathbf{y_1}$, $\mathbf{y_2}$ from class B, our method finds the optimal hard negatives by calculating the minimum distance between the curves joining these points.}
    \label{fig1}
\end{figure}
\par To train the deep networks, several loss functions with desirable properties have been formulated, which enable the neural network to learn the mapping from the data space to the embedding space. Conventional methods, such as contrastive loss \cite{cont1,cont2} which takes tuples with two samples and triplet loss \cite{triplet1,triplet2} which takes three sample tuples, consider the similarity between few samples. In contrast, methods like lifted structure loss \cite{lift-struct} aim to exploit all the samples present in a batch to learn more informative representations. Other methods \cite{hist,min1} that utilize several samples in the loss have also been proposed.
\par However, even when considering the whole batch, not all the samples are able to contribute to the loss term. This is because many of them already satisfy the constraints present in the loss. As a result, these samples are not sufficiently informative and lead to low gradient values. To overcome this problem, the idea of using hard samples has been considered, \textit{i.e.} positive samples which lie far away and negative samples which lie closer. Several works on hard negative mining \cite{min1,min2,min3,min4,triplet1} have been proposed. Mining-based strategies generally look for the most informative samples in the dataset and are prone to learning biased mappings, which do not generalize well to the entire dataset. 
On the other hand, hard negative generation-based methods \cite{daml,hdml,htg} utilize an additional sub-network, either an autoencoder or a generative adversarial network \cite{gan} as the generator. This can result in harder optimization \cite{wgan} and an increase in training time and computation.
\par To address the limitations of existing hard negative mining and generation methods, we propose a novel approach, which looks for optimal hard negative samples (LoOp) in the embedding space. This is done by finding points that minimize the distance between the curves joining a pair of positives and a pair of negatives, as shown in Fig \ref{fig1}. The curve joining a pair of points belonging to the same class lies in the region belonging to that class. Hence, finding the minimum distance with another curve, which is representative of some other class, allows us to consider the most informative pair of samples for computing the loss, as described in Section \ref{prob}. Unlike mining-based methods, our approach does not neglect any samples, and unlike generation-based methods, it does not increase the training complexity or optimization difficulty. It can be easily integrated with various metric learning-based losses. We include several experimental results in Section \ref{exps} to demonstrate the efficacy of our approach.

\par \textbf{Contributions} We propose a novel approach, LoOp, which finds optimal hard negatives in the embedding space, 
and maximizes the contribution of each tuple in computing the pair-based metric learning loss. Our approach presents the solution to a general problem, namely finding the minimum distance between two bounded curves, which can be useful in other applications as well. Generating hard negatives in the embedding space, our approach utilizes all training samples and does not rely on a subset of samples like mining-based methods. It also avoids the computational load and training complexity introduced by generation-based methods. We also explore the optimality of LoOp with a gradient-based theoretical analysis of the loss function. Our approach generalizes well to various metric learning losses, and it can be easily combined without any increase in optimization difficulty or additional parameters. It outperforms state-of-the-art mining and generation-based methods on three benchmark datasets, Cars196 \cite{cars196}, CUB-200-2011 \cite{cub200}, and Stanford Online Products \cite{lift-struct}.

\vspace{-2mm}
\section{Related Work}
\textbf{Hard Negative Mining} The samples which satisfy the criteria used for training do not contribute to the loss and lead to smaller gradients and slower convergence. To combat this issue, several sampling and mining strategies \cite{min1,min2,min3,min4,triplet1} have been developed, which use hard positives and negatives for training. Schroff \textit{et al.} \cite{triplet1} propose `semi-hard' triplet mining, where the negatives, which lie close to the anchor but farther than the positives, are selected. Harwood \textit{et al.} \cite{min2} present a smart mining-based method that adaptively selects the most informative samples. The drawback of such methods is that the trained models overfit on a subset of hard samples and underfit on the `easy' samples. As they rely on a subset of training samples, they fail to exploit the information other samples provide. 
\par \textbf{Hard Negative Generation} In contrast to mining-based methods, generation-based methods \cite{daml,hdml,htg} try to exploit all the training samples by extracting the semantic information required to generate synthetic samples which act as hard negatives. Deep adversarial metric learning (DAML) \cite{daml} trains the embedding network and the hard negative generator network in an adversarial manner to generate hard negatives. In hardness-aware deep metric learning (HDML) \cite{hdml}, synthetic samples are created via interpolation in the embedding space. This is followed by finding the corresponding label-preserving mappings in the feature space using an autoencoder. They also control the level of hardness while training. Although these methods utilize all the training samples, they require an additional network as the generator. This increases the training time and computational load and can also lead to optimization difficulty \cite{wgan}.
\par \textbf{Embedding Space Augmentation} Some methods \cite{ftl,embex} augment the embedding space directly to obtain useful synthetic samples. Yin \textit{et al.} \cite{ftl} assume that all classes follow a Gaussian distribution and translate samples from different classes about their means to generate new ones. Embedding expansion (EE) \cite{embex} is a linear interpolation-based method to generate synthetic samples and mine for hard negatives by considering the pair-wise distances between real and synthetic points of different classes. It exhibits a trade-off between the quality of hard negatives and the computational load of calculating the pair-wise distances. 

\par Other works try to supplement the metric learning losses by introducing a regularization to optimize the direction of displacement of samples \cite{right_dir} or maximize the spread-out of feature descriptors \cite{spread_out}. Another work introduced second-order similarity \cite{sosnet} inspired by graph matching and clustering as a regularizer for learning local feature descriptors.  
\vspace{-4mm}
\section{Methodology}
\label{method}
Let $\mathcal{Z}$ denote the set of data points, $\mathcal{X}$ denote the set of embeddings, and $h(\cdot;\theta): \mathcal{Z}\xrightarrow{}\mathcal{X}$ denote the mapping learnt by the neural network $h$ with parameters $\theta$. $\mathcal{C}$ denotes the set of classes to which the data points belong. Let $\mathbf{Z}$ denote the array of data points sampled for training. For any index $i$, $\mathbf{Z}[i]$ denotes the data sample, $\mathbf{X}[i]$ denotes the corresponding embedding, and $\mathbf{c}[i]$ denotes its class. 
\vspace{-1mm}
\subsection{Problem Formulation}
\label{prob}
\par Consider two 
 pairs of points in the embedding space: $\mathbf{x_1}$, $\mathbf{x_2}$ and $\mathbf{y_1}$, $\mathbf{y_2}$, belonging to two different classes. They are $l_2$-normalized, and lie on a hypersphere of unit radius. Most pair-wise metric learning-based losses try to consider the gap between $d(\mathbf{x_1},\mathbf{x_2})$ and $d(\mathbf{x_1},\mathbf{y_1})$ in some way,
where $d$ denotes the Euclidean distance. To consider informative samples which allow for larger values of the loss function, we can try to increase this gap. As $d(\mathbf{x_1},\mathbf{x_2})$ represents the distance between two positives, it utilizes both the samples from that class available in the batch and although we can try to generate $\mathbf{x_3}$ such that $d(\mathbf{x_1},\mathbf{x_3})>d(\mathbf{x_1},\mathbf{x_2})$, it would be mathematically difficult to ensure its class belongingness without more information (or assumptions). On the other hand, we can minimize the distance between the negatives represented by $d(\mathbf{x_1},\mathbf{y_1})$ by utilizing the remaining samples ($\mathbf{x_2}$, $\mathbf{y_2}$). In other words, we can find two points, $\mathbf{p_1}$ and $\mathbf{p_2}$, which lie on the geodesic curves joining $\mathbf{x_1}$ and $\mathbf{x_2}$, and $\mathbf{y_1}$ and $\mathbf{y_2}$, respectively, such that
\begin{align}
\label{dis_eqn}
    d(\mathbf{p_1},\mathbf{p_2}) = ||\mathbf{p_1}-\mathbf{p_2}||_2
     =\sqrt{2(1-\mathbf{p_1}\cdot \mathbf{p_2})},
\end{align}
is minimized. 
This is visualized in Fig. \ref{fig1}. A toy example demonstrating the behavior of triplet loss using different sets of $\mathbf{p_1}$, $\mathbf{p_2}$ 
and a gradient-based analysis is included in the supplementary.
\par \textbf{Assumption on Data Distribution} We assume that data distributions of different classes lying on the hypersphere are spherical-homoscedastic \cite{sph-homo}, defined as follows.
\vspace{-1mm}
\begin{customthm}{1}\label{def1}
Two distributions are said to be spherical-homoscedastic if their covariances have identical eigenvalues, i.e. the distributions are identically shaped.
\end{customthm}
\vspace{-2mm}
\begin{customprop}{1}\label{prop1}
If the pairs of points $\mathbf{x_1}$, $\mathbf{x_2}$ and $\mathbf{y_1}$, $\mathbf{y_2}$ (from two different classes) belong to spherical-homoscedastic distributions, then the points on the curves $\arc{\mathbf{x_1x_2}}$ and $\arc{\mathbf{y_1y_2}}$ have a higher probability of belonging to the same classes as $\mathbf{x_1}$, $\mathbf{x_2}$ and $\mathbf{y_1}$, $\mathbf{y_2}$, respectively, than the other classes.
\end{customprop}
\vspace{-1mm}
Experimental validation for the assumption on data distribution (Definition \ref{def1}) in our pipeline and the proof of Proposition \ref{prop1} can be found in the supplementary material.
\par $\mathbf{p_1}$ is obtained by rotating $\mathbf{x_1}$ towards $\mathbf{x_2}$ along the curve $\arc{\mathbf{x_1x_2}}$ by angle $\alpha$, which lies between 0 and $\alpha_0=\cos^{-1}(\mathbf{x_1}\cdot \mathbf{x_2})$. Similarly, $\mathbf{p_2}$ is obtained by rotating $\mathbf{y_1}$ towards $\mathbf{y_2}$ along $\arc{\mathbf{y_1y_2}}$ by angle $\beta$, which lies between 0 and $\beta_0=\cos^{-1}(\mathbf{y_1}\cdot \mathbf{y_2})$. This is done in three steps. First, the basis vectors for the space spanned by $\mathbf{x_1}$ and $\mathbf{x_2}$ are calculated using Gram-Schmidt orthogonalization, \textit{i.e.}
\begin{equation}
\vspace{-1mm}
    \mathbf{n_1}=\mathbf{x_1} \text{ ; } \mathbf{n_2}=\frac{\mathbf{x_2}-(\mathbf{x_1}\cdot \mathbf{x_2})\mathbf{x_1}}{||\mathbf{x_2}-(\mathbf{x_1}\cdot \mathbf{x_2})\mathbf{x_1}||_2}.
\end{equation} 
\vspace{-1mm}
Second, the rotation matrix $\mathbf{R}$ is calculated using Rodriguez theorem, as follows:
\begin{align*}
\vspace{-1mm}
    \mathbf{R}&=\mathbf{I}+\sin\alpha(\mathbf{n_2}\mathbf{n_1}^T-\mathbf{n_1}\mathbf{n_2}^T)\\
    &-(1-\cos\alpha)(\mathbf{n_1}\mathbf{n_1}^T+\mathbf{n_2}\mathbf{n_2}^T),
    \vspace{-1mm}
\end{align*}
where $\mathbf{I}$ is the identity matrix and $(\cdot)^T$ is the transpose operator. Finally, we find $\mathbf{p_1}=\mathbf{R}\mathbf{x_1}$. Simplifying, we get:
\begin{equation}
\label{p1eqn}
\vspace{-1mm}
    \mathbf{p_1}=\mathbf{n_1}\cos\alpha+\mathbf{n_2}\sin\alpha.
    \vspace{-1mm}
\end{equation}
Similarly, $\mathbf{p_2}$ is obtained as:
\begin{equation}
\label{p2eqn}
\vspace{-1mm}
    \mathbf{p_2}=\mathbf{n_3}\cos\beta+\mathbf{n_4}\sin\beta,
    \vspace{-1mm}
\end{equation}
where $\mathbf{n_3}$ and $\mathbf{n_4}$ are the basis vectors obtained by Gram-Schmidt orthogonalization for $\mathbf{y_1}$ and $\mathbf{y_2}$.
\par Incorporating these expressions for $\mathbf{p_1}$ and $\mathbf{p_2}$ in (\ref{dis_eqn}) and simplifying
, the objective $f$ to be minimized, is given by:
\begin{align}
\vspace{-1mm}
    f&(\alpha,\beta)=-\mathbf{p_1}\cdot\mathbf{p_2}=a\sin\alpha\sin\beta\nonumber\\&+b\cos\alpha\sin\beta+c\sin\alpha\cos\beta+d\cos\alpha\cos\beta,
\vspace{-1mm}
\end{align}
where $a=-\mathbf{n_2}\cdot \mathbf{n_4}$, $b=-\mathbf{n_1}\cdot \mathbf{n_4}$, $c=-\mathbf{n_2}\cdot \mathbf{n_3}$, and $d=-\mathbf{n_1}\cdot \mathbf{n_3}$.
\begin{table*}[t]
    \centering
    \large
    \resizebox{1.97\columnwidth}{!}{\begin{tabular}{ccccccc}
    \toprule
    \rule{0pt}{9pt} Case & $\alpha$ & $\beta$ & $\lambda_1$ & $\lambda_2$ & $\lambda_3$ & $\lambda_4$ \\
    \midrule
    \rule{0pt}{12pt} 1 & 0 & $\tan^{-1}\left(\frac{b}{d}\right)$
         & $-\frac{\partial f}{\partial \alpha}\big|_{\alpha=0,\beta=\hat\beta}$
         & 0 & 0 & 0\\
         
    \rule{0pt}{12pt} 2 & $\alpha_0$ & $\tan^{-1}\left(\frac{a\sin\alpha_0+b\cos\alpha_0}{c\sin\alpha_0+d\cos\alpha_0}\right)$ 
         & 0 & $\frac{\partial f}{\partial \alpha}\big|_{\alpha=\alpha_0, \beta=\hat\beta}$ 
         & 0 & 0\\
         
    \rule{0pt}{12pt} 3 & $\tan^{-1}\left(\frac{c}{d}\right)$ & 0
         & 0 & 0 & $-\frac{\partial f}{\partial \beta}\big|_{\alpha=\hat\alpha,\beta=0}$ 
         & 0\\
         
    \rule{0pt}{12pt} 4 & $\tan^{-1}\left(\frac{a\sin\beta_0+c\cos\beta_0}{b\sin\beta_0+d\cos\beta_0}\right)$ & $\beta_0$ 
         & 0 & 0 & 0 & $\frac{\partial f}{\partial \beta}\big|_{\alpha=\hat\alpha,\beta=\beta_0}$\\
         
    \rule{0pt}{12pt} 5 & 0 & 0
         & $-\frac{\partial f}{\partial \alpha}\big|_{\alpha=0,\beta=0}$ & 0 & $-\frac{\partial f}{\partial \beta}\big|_{\alpha=0,\beta=0}$ 
          & 0\\
          
    \rule{0pt}{12pt} 6 & 0 & $\beta_0$
         & $-\frac{\partial f}{\partial \alpha}\big|_{\alpha=0,\beta=\beta_0}$
         & 0 & 0 & 
         $\frac{\partial f}{\partial \beta}\big|_{\alpha=0,\beta=\beta_0}$
         \\
         
    \rule{0pt}{12pt} 7 & $\alpha_0$ & 0
         & 0 & 
         $\frac{\partial f}{\partial \alpha}\big|_{\alpha=\alpha_0,\beta=0}$ & $-\frac{\partial f}{\partial \beta}\big|_{\alpha=\alpha_0,\beta=0}$
         & 0\\
         
    \rule{0pt}{12pt} 8 & $\alpha_0$ & $\beta_0$
         & 0 & $\frac{\partial f}{\partial \alpha}\big|_{\alpha=\alpha_0,\beta=\beta_0}$ & 0 & $\frac{\partial f}{\partial \beta}\big|_{\alpha=\alpha_0,\beta=\beta_0}$\\
    \bottomrule
    \end{tabular}}
    \caption{Expressions for $\alpha$, $\beta$, and the KKT multipliers for 8 cases of complementary slackness.}
    \label{tab:kkt_cases}
\end{table*}

\par Further, we need to consider two constraints on $\alpha$, so that $\mathbf{p_1}$ remains between 
$\mathbf{x_1}$ and $\mathbf{x_2}$. 
They are given as:
\begin{equation}
\vspace{-1mm}
    g_1=-\alpha\leq0 \text{ ; } g_2=\alpha-\alpha_0\leq0.
    \vspace{-1mm}
\end{equation}

 Similarly, the constraints on $\beta$ are given as:
\begin{equation}
\vspace{-1mm}
    g_3=-\beta\leq0 \text{ ; }
    g_4=\beta-\beta_0\leq0.
\end{equation}
 The Lagrangian function for the constrained optimization problem is given by:
\begin{equation}
\label{lag}
\vspace{-1mm}
    L(\alpha,\beta,\lambda_1,\lambda_2,\lambda_3,\lambda_4)=f(\alpha,\beta)-\sum_{i=1}^{4}\lambda_ig_i,
    \vspace{-1mm}
\end{equation}

where $\lambda_i$, $i=1,2,3,4$ are Karush-Kuhn-Tucker (KKT) \cite{karush,kt} multipliers.

\subsection{Finding Optimal Distance}
\label{solution_section}
\par As the constraints are differentiable without any critical points, we consider the KKT conditions \cite{karush,kt} to obtain the solution for this problem. They are listed as follows:
\begin{align}
\label{delLdelalpha}
\frac{\partial L}{\partial \alpha}&=a \cos\alpha\sin\beta-b\sin\alpha\sin\beta+c\cos\alpha\cos\beta\nonumber\\
&-d\sin\alpha\cos\beta+\lambda_1-\lambda_2=0,\\
\label{delLdelbeta}
\frac{\partial L}{\partial \beta}&=a \sin\alpha\cos\beta+b\cos\alpha\cos\beta-c\sin\alpha\sin\beta\nonumber\\
&-d\cos\alpha\sin\beta+\lambda_3-\lambda_4=0,
\end{align}
\vspace{-5mm}
\begin{align}
\label{comp_slack}
\lambda_ig_i&=0 \text{ ; } i=1,2,3,4,\\
\label{lambdas0}
\lambda_i&\leq 0 \text{ ; } i=1,2,3,4,\\
\label{gs0}
g_i&\leq 0 \text{ ; } i=1,2,3,4.
\end{align}
\vspace{-1mm}
\begin{figure}[h]
    \centering
    \includegraphics[scale=0.32]{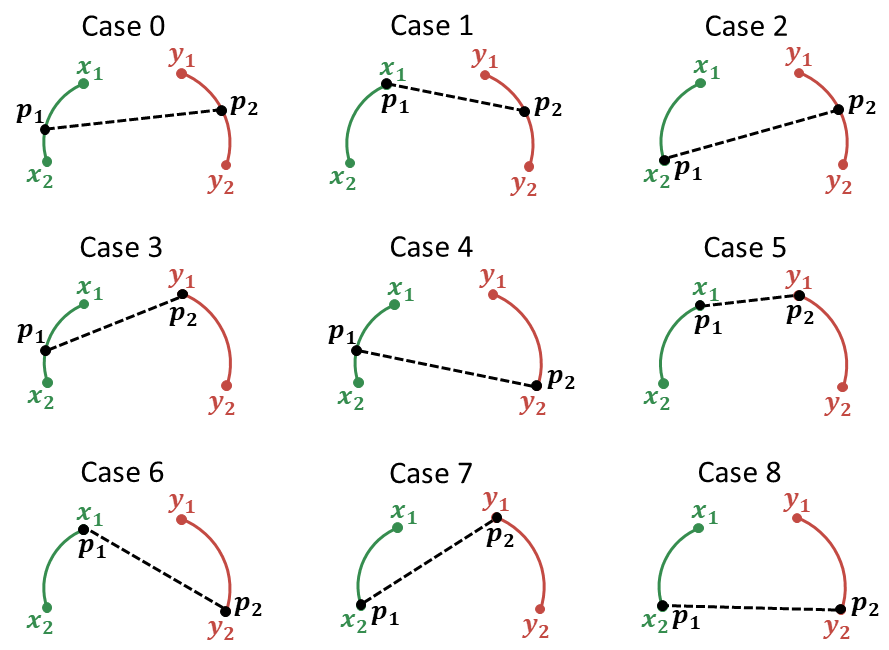}
    \caption{Illustrations of the 9 cases in which the KKT conditions for minimizing the distance between $\arc{\mathbf{x_1x_2}}$ and $\arc{\mathbf{y_1y_2}}$ can be satisfied. In case 0, $\mathbf{p_1}$ and $\mathbf{p_2}$ lie between the end points of the respective curves. In cases 1-4, only one of them is an end point of the respective curve. In cases 5-8, both of them are end points of the respective curves.}
    \label{cases_fig}
    \vspace{-4mm}
\end{figure}
\par There are 16 possible ways to satisfy (\ref{comp_slack}). However, some constraints cannot be binding simultaneously (e.g. if $g_1=0$, $g_2\neq0$), which leaves 9 cases to consider, which are shown in Fig. \ref{cases_fig}. The solutions are discussed below in brief. The proofs can be found in the supplementary material.
\par We first consider the case where none of the constraints is binding (Case 0). In order to satisfy (\ref{comp_slack}), $\lambda_i=0$ for $i=1,2,3,4$. In this case, (\ref{lambdas0}) is also satisfied. Leveraging (\ref{delLdelalpha}) and (\ref{delLdelbeta}), we get a quadratic equation in $\tan\alpha$. On solving it and finding the corresponding value of $\beta$ using (\ref{delLdelalpha}) or (\ref{delLdelbeta}), we get two sets of solutions:
\vspace{-2mm}
\begin{align*}
    \hat\alpha=\tan^{-1}\left(\frac{A\pm\sqrt{A^2+4(ab+cd)^2}}{2(ab+cd)}\right),\\
    \hat\beta=\tan^{-1}\left(\frac{B\mp\sqrt{B^2+4(ac+bd)^2}}{2(ac+bd)}\right),
\end{align*}
where $A=a^2-b^2+c^2-d^2$ and $B=a^2+b^2-c^2-d^2$. The obtained $\hat\alpha$ and $\hat\beta$ are used to verify the validity of (\ref{gs0}). 
\par Similarly, the KKT conditions are checked for the remaining 8 cases as well. The expressions for $\alpha$, $\beta$, and the KKT multipliers for various cases are listed in Table \ref{tab:kkt_cases}. The first 4 rows of the table cover the cases where one of the optimal points is an end point of a curve, while the other can be any point on the other curve. The next 4 rows of the table cover the cases where both the optimal points are one of the two end points. The values obtained using these expressions are used to verify the conditions in (\ref{lambdas0}) and (\ref{gs0}). In case any condition is violated, the solution is discarded. 
\par Once $\hat\alpha$ and $\hat\beta$ are obtained, they are substituted in (\ref{p1eqn}) and (\ref{p2eqn}) to obtain $\mathbf{p_1}$ and $\mathbf{p_2}$. Finally, $d$ is calculated using (\ref{dis_eqn}), which is used in different metric learning losses, as discussed below. 
\subsection{Optimal Hard Negative Embeddings for Deep Metric Learning}
\par Let $d_{i,j}$ denote the distance between $\mathbf{X}[i]$ and $\mathbf{X}[j]$ and $s_{i,j}$ denote the similarity, given by $s_{i,j}=1-\frac{d_{i,j}^2}{2}$. Let $d_{i,j,k,l}$ denote the optimal distance between the curves joining $\mathbf{X}[i]$ and $\mathbf{X}[j]$, and $\mathbf{X}[k]$ and $\mathbf{X}[l]$, such that $\mathbf{c}[i]=\mathbf{c}[j]\neq\mathbf{c}[k]=\mathbf{c}[l]$. 
We evaluate our method for the following metric learning losses:
\vspace{-3mm}
\subsubsection{Triplet Loss}
\par One of the earliest metric learning losses, triplet loss \cite{triplet1,triplet2} tries to pull a positive sample belonging to the same class as the anchor, close to the anchor, and push a negative sample belonging to a different class away from the anchor, by a margin $m$. It is formulated as follows: 
\vspace{-1mm}
\begin{align}
    \mathcal{L}_{Tri}=\frac{1}{|\mathcal{P}|}\mathop{\sum_{(i,j)\in\mathcal{P}}}_{k:\mathbf{c}[i]\neq\mathbf{c}[k]}\left[d_{i,j}-d_{i,k}+m\right]_+,
\end{align}
where $[\cdot]_+$ denotes the hinge function, and $\mathcal{P}$ denotes the set of classes in the batch. 
The modified triplet loss, obtained by incorporating LoOp, is given by:
\begin{align}
    \mathcal{L}'_{Tri}=\frac{1}{|\mathcal{P}|}\mathop{\sum_{(i,j)\in\mathcal{P}}}_{k,l:\mathbf{c}[i]\neq\mathbf{c}[k]=\mathbf{c}[l]}\left[d_{i,j}-d_{i,j,k,l}+m\right]_+.
\end{align}
\vspace{-2mm}
\subsubsection{HPHN Triplet Loss}
\par In order to better utilize the samples present in a batch, hard positive and hard negative (HPHN) mining \cite{hphn1,hphn2} has been proposed. The expression for HPHN-triplet loss is given by:
\begin{align*}
    \mathcal{L}_{HPHNtri}=\frac{1}{|\mathcal{P}|}\sum_{(i,j)\in\mathcal{P}}\bigg[\max\left(\max_{\mathbf{c}[i]= \mathbf{c}[k]}d_{i,k},\max_{\mathbf{c}[j]= \mathbf{c}[l]}d_{j,l}\right)\nonumber\\
    +m-\min\left(\min_{\mathbf{c}[i]\neq \mathbf{c}[k]}d_{i,k},\min_{\mathbf{c}[j]\neq \mathbf{c}[l]}d_{j,l}\right)\bigg]_+.
\end{align*}
The modified HPHN-triplet loss, obtained by incorporating LoOp, is given by:
\begin{align*}
    \mathcal{L}'_{HPHNtri}=\frac{1}{|\mathcal{P}|}\sum_{(i,j)\in\mathcal{P}}\bigg[\max\left(\max_{\mathbf{c}[i]= \mathbf{c}[k]}d_{i,k},\max_{\mathbf{c}[j]= \mathbf{c}[l]}d_{j,l}\right)\nonumber\\
    +m-\min_{\mathbf{c}[i]\neq \mathbf{c}[k]=\mathbf{c}[l]}d_{i,j,k,l}\bigg]_+.
\end{align*}
\vspace{-2mm}
\subsubsection{Lifted Structure Loss}
\par The lifted structure (LS) loss \cite{lift-struct} tries to push a pair of samples belonging to ane class away from all the samples belonging to some different class. It is formulated as:
\begin{align}
    \mathcal{L}_{LS}=\frac{1}{|\mathcal{P}|}\sum_{(i,j)\in\mathcal{P}}\bigg[d_{i,j}+m-\nonumber\\
    \min\left(\min_{\mathbf{c}[i]\neq \mathbf{c}[k]}d_{i,k},\min_{\mathbf{c}[j]\neq \mathbf{c}[l]}d_{j,l}\right)\bigg]_+.
\end{align}
\par When only 2 samples of each class are present in the batch, the HPHN-triplet loss becomes the same as LS loss. The modified LS loss, obtained by incorporating LoOp, is given by:
\begin{align*}
    \mathcal{L}'_{LS}=\frac{1}{|\mathcal{P}|}\sum_{(i,j)\in\mathcal{P}}\bigg[d_{i,j}+m-\min_{\mathbf{c}[i]\neq \mathbf{c}[k]=\mathbf{c}[l]}d_{i,j,k,l}\bigg]_+.
\end{align*}
\vspace{-2mm}
\subsubsection{Multi-Similarity Loss}
\par The multi-similarity (MS) loss \cite{msloss} considers self-similarity and relative similarities between positive as well as negative pairs to learn the embedding space. It involves mining for hard negative and positive pairs, where a pair of negatives $(\mathbf{X}[i],\mathbf{X}[j])$ (for $\mathbf{c}[i]\neq \mathbf{c}[j]$) is selected when: 
\begin{equation*}
    s^-_{i,j}>\min_{\mathbf{c}[i]= \mathbf{c}[k]}s_{i,k}-\epsilon.
\end{equation*}
Similarly, a pair of positives is selected, when:
\begin{equation*}
    s^+_{i,j}<\max_{\mathbf{c}[i]\neq \mathbf{c}[k]}s_{i,k}+\epsilon.
\end{equation*} 
\par In both the conditions, $\epsilon$ is a hyperparameter. The second step is general pair weighting, and the final expression for MS loss is given by:
\begin{align*}
    \mathcal{L}_{MS}=\frac{1}{|\mathcal{P}|}\sum_{i\in\mathcal{P}}\bigg[\frac{1}{\alpha_m}log\Big(1+\sum_{\mathbf{c}[i]=\mathbf{c}[j]}e^{-\alpha_m(s_{i,j}-\lambda)}\Big)\nonumber\\+\frac{1}{\beta_m}log\Big(1+\sum_{\mathbf{c}[i]\neq \mathbf{c}[j]}e^{\beta_m(s_{i,j}-\lambda)}\Big)\bigg],
\end{align*}
where $\alpha_m$ and $\beta_m$ are hyperparameters, and $\lambda$ denotes the margin. For the modified version of MS loss, obtained by incorporating LoOp, the criteria for selecting hard negatives is given as follows:
\begin{equation*}
    s^-_{i,k,j,l}>\min_{\mathbf{c}[i]= \mathbf{c}[k]}s_{i,k}-\epsilon.
\end{equation*}
The expressions for selecting hard positives as well as for general pair weighting remain unchanged.
\subsection{Implementation}
\par Algorithm \ref{alg:o} presents the steps involved in the proposed approach. In order to make the implementation efficient, the calculations for all possible combinations are carried out in parallel. Given a batch size $B_S$ and number of samples per class $N$, the number of such combinations $C$ is given as:
\begin{align*}
    C=\frac{\frac{B_S}{2}(\frac{B_S}{2}-1)}{2}-\frac{B_S}{N}\times\frac{\frac{N}{2}(\frac{N}{2}-1)}{2}=\frac{B_S(B_S-N)}{8}.
\end{align*}
The runtime of our method is documented in Section \ref{time}.
\vspace{-1mm}
\begin{algorithm}
\caption{Training with optimal hard negatives}
\label{alg:o}
\SetKwInOut{Input}{input}\SetKwInOut{Output}{output}
\SetKw{KwRet}{return}\SetKw{Return}{return}
\Input{Set of training images and labels, ensuring number of samples of each class is even; hyperparamters depending on the loss function; number of iterations $T$}
\Output{Trained network parameters $\theta$}
\BlankLine
Initialize $\theta$\\
\For{\emph{iteration}=1,2,...,T}{
Extract embeddings of the images in the training batch and perform $l_2$-normalization\\
Create pairs of embeddings (by choosing alternate elements from $\mathbf{X}$) to get the sets $\mathbf{X_1}$ and $\mathbf{X_2}$, with the corresponding classes $\mathbf{c_{X_1}}=\mathbf{c_{X_2}}$\\
Create 4 arrays: $\mathbf{X_1}$, $\mathbf{X_2}$, $\mathbf{Y_1}$, $\mathbf{Y_2}$, containing the possible combinations of the aforementioned pairs, ensuring that $\mathbf{c_{X_1}}\neq\mathbf{c_{Y_1}}$\\
Calculate the optimal distance between each combination of pairs, \textit{i.e.} between the corresponding elements of the arrays $\mathbf{X_1}$, $\mathbf{X_2}$, $\mathbf{Y_1}$, $\mathbf{Y_2}$ in parallel\\ 
Find the minimum distance for each unique pair in $\mathbf{X_1}$, $\mathbf{X_2}$ \\
Calculate the metric learning-based loss using the set of optimal distances\\
Update network parameters
}
\Return{$\theta$}
\end{algorithm}
\vspace{-2mm}
\section{Experiments}
\label{exps}
We conduct numerous experiments to evaluate the effectiveness of LoOp for both image clustering and retrieval tasks \cite{lift-struct}. 
We report the standardized metrics $F_1$ score and normalized mutual information (NMI) for image clustering \cite{ret1}, and Recall@K values for image retrieval.   
\par \textbf{Datasets} We evaluate our method on three widely used benchmark datasets described below. We use the conventional zero-shot setting \cite{lift-struct, daml}, \textit{i.e.} no intersection in train and test classes. Results for train-validate-test split are included in the supplementary.
\vspace{-1mm}
\begin{itemize}
    \item The CUB-200-2011 dataset \cite{cub200} comprises 11,788 images of 200 bird species. We use the first 100 species (5864 images) for training and the remaining 100 (5924 images) for testing. 
    \vspace{-1mm}
    \item The Cars196 dataset \cite{cars196} consists of 16,185 images of 196 classes of cars. We use the first 98 car classes (8054 images) as the training data and the remaining 98 classes (8131 images) as testing data. 
    \vspace{-1mm}
    \item The Stanford Online Products (SOP) dataset \cite{lift-struct} contains 120,053 images of 22,634 online products listed on eBay.com. The train set consists of 59,551 images of first 11,318 product classes, while the test set uses the remaining 60,502 images of 11,316 classes.
\end{itemize}

\par \textbf{Settings} We use the MXNet package for implementing our proposed approach. All the experiments were conducted using one NVIDIA GeForce RTX 2080 Ti, 11 GB memory. For pre-processing input images, we resize them to $256\times256$, randomly crop them to size $227\times227$, horizontally flip them left or right. We use an ImageNet ILSVRC \cite{imagenet} pre-trained GoogLeNet \cite{googlenet} with a fully connected randomly initialized layer as our feature extracting network. For a fair comparison, we compare with methods that use the same network. Results and comparisons on other network architectures have been included in the supplementary. Throughout the experiments, we use 512-dimensional feature embedding vectors. We use the Adam optimizer \cite{adam} for training our models, with $\beta_1=0.9, \beta_2=0.999$, learning rate of $10^{-4}$. The hyperparameters $\alpha_m$ and $\beta_m$ are set as 2 and 50, respectively. For Cars196, CUB-200-2011 datasets, we use a batch size of 32, and for the SOP dataset, we set the batch size to 128. The SOTA methods use a larger batch size of 120 or 128 for the Cars196, CUB-200-2011 dataset as well. As shown in Section \ref{param_effect}, larger batch size provides a boost in performance metrics. t-SNE visualizations at different epochs demonstrating the training process can be found in the supplementary.
\subsection{Effect of Hyperparameters}
\label{param_effect}
\par In this section, we study the effects of changing the values of various hyperparameters on the Recall@1 (\%) performance. 
\begin{figure}[h!]
    \centering
    \begin{subfigure}[b]{0.48\columnwidth}
         \centering
         \includegraphics[scale=0.185]{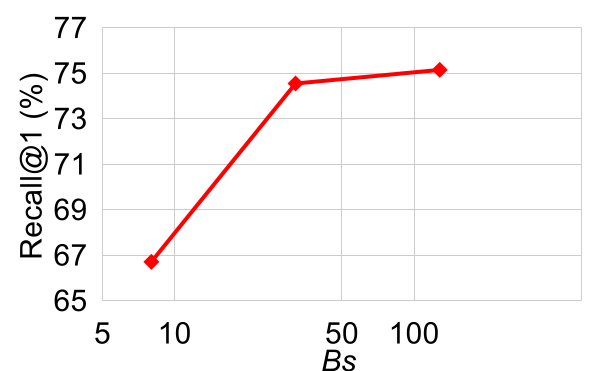}
    \caption{Effect of $B_S$.}
    \label{bs}
     \end{subfigure}
     \hfill
    \begin{subfigure}[b]{0.48\columnwidth}
         \centering
         \includegraphics[scale=0.185]{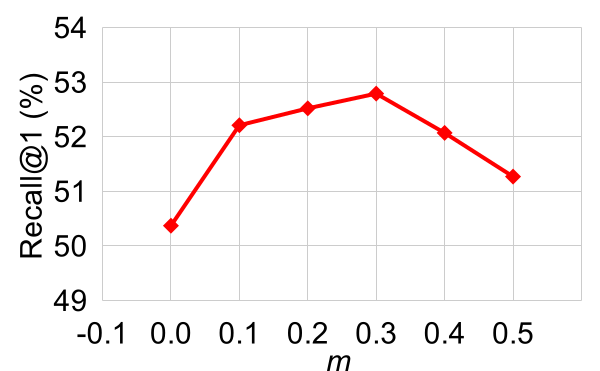}
    \caption{Effect of $m$.}
    \label{m}
     \end{subfigure}\\
    \begin{subfigure}[b]{0.48\columnwidth}
         \centering
         \includegraphics[scale=0.185]{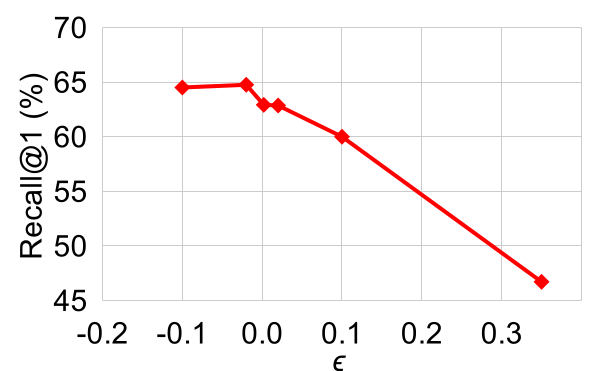}
    \caption{Effect of $\epsilon$.}
    \label{eps}
     \end{subfigure}
     \hfill
     \begin{subfigure}[b]{0.48\columnwidth}
         \centering
         \includegraphics[scale=0.185]{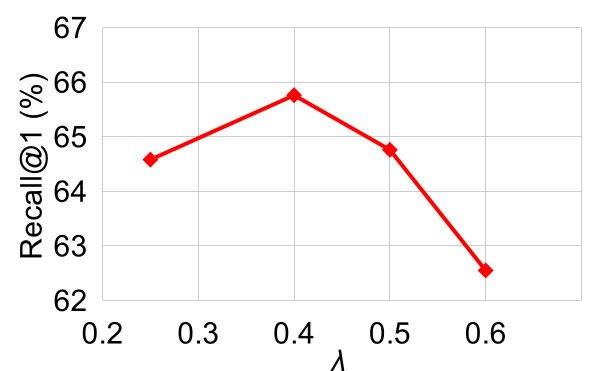}
    \caption{Effect of $\lambda$.}
    \label{lam}
     \end{subfigure}
\caption{Variation in Recall@1 (\%) values with various hyperparameters.}
\vspace{-2mm}
\end{figure}
\par \textbf{Effect of Batch Size} 
Fig. \ref{bs} shows the effect of varying $B_S$ (log scale is used for $x$-axis) for Cars196 dataset using HPHN-triplet loss. The number of unique classes in the batch is 4. It is observed that as $B_S$ increases, there is an increase in the Recall@1 value. However, the rate of increase is lower for higher $B_S$.
\par \textbf{Effect of Margin} 
Fig. \ref{m} shows the effect of varying $m$, which is used in triplet, HPHN-triplet, and LS losses. These values are for CUB-200-2011 dataset for LS loss, with $B_S=32$. It is observed that as $m$ increases, the performance initially increases, and then drops. This shows that for smaller values of $m$, the embeddings of dissimilar classes are not sufficiently separated, whereas for higher values of $m$, they are forced to form more compact and distant clusters. This might lead to overfitting resulting in subsequent drop in performance.
\par \textbf{Effect of $\epsilon$ and $\lambda$} Next, we consider the effect of varying two hyperparameters of the MS loss. In both cases, the results are reported for the Cars196 dataset, with $B_S=8$ and $N=4$. Fig. \ref{eps} presents the results of varying $\epsilon$, for $\lambda=0.5$. From the range of the $y$-axis, it can be observed that the performance of MS loss is quite sensitive to this parameter. It is noteworthy that since our approach finds the hardest negatives for each pair in a batch, the role of $\epsilon$ (to make the criteria for selection of hard positives more strict) is diminished, and we get performance gains for lower values of $\epsilon$. Fig. \ref{lam} presents the results of varying $\lambda$, for $\epsilon=-0.02$. As the value of similarity lies between $-1$ and $1$, embeddings belonging to the same class are constrained to have similarity more than $\lambda$, while those belonging to dissimilar classes should have a smaller similarity than $\lambda$. Varying $\lambda$ for MS loss has a similar effect as varying $m$ for LS loss. 
\vspace{-1mm}
\subsection{Effect of $l_2$-normalization}
\par Most of the metric learning losses considered in this work perform $l_2$-normalization on the embeddings before computing the loss. In order to observe the effect of $l_2$-normalization, we consider the case where the optimal points lie on the line segments $\overline{\mathbf{x_1x_2}}$ and $\overline{\mathbf{y_1y_2}}$ rather than the curves joining these points. In other words, the optimal points are not constrained to lie on the hypersphere. The optimization problem is then formulated differently. A brief discussion is presented here, and the details can be found in the supplementary material.
\par The optimal points $\mathbf{p_1}$ and $\mathbf{p_2}$ can be obtained as:
\begin{align}
    \mathbf{p_1}&=(1-k_1)\mathbf{x_1}+k_1\mathbf{x_2},\nonumber\\
    \mathbf{p_2}&=(1-k_2)\mathbf{y_1}+k_2\mathbf{y_2},
\end{align}
where $k_1$ and $k_2$ are the parameters to be optimized. $f$ can be obtained by incorporating these expressions in (\ref{dis_eqn}). The constraints in this case are given as:
\begin{align}
    g_1&=-k_1\leq0 \text{ ; } g_2=k_1-1\leq0,\nonumber\\
    g_3&=-k_2\leq0 \text{ ; } g_4=k_2-1\leq0.
\end{align}
\par The Lagrangian function can be obtained using (\ref{lag}). The partial derivatives are given as:
\begin{align}
    \frac{\partial L}{\partial k_1}=&(\mathbf{x_2}-\mathbf{x_1})\cdot(k_1(\mathbf{x_2}-\mathbf{x_1})-k_2(\mathbf{y_2}-\mathbf{y_1})\nonumber\\+&\mathbf{x_1}-\mathbf{y_1})+\lambda_1-\lambda_2=0,\\
   \frac{\partial L}{\partial k_2}=&(\mathbf{y_2}-\mathbf{y_1})\cdot(-k_1(\mathbf{x_2}-\mathbf{x_1})+k_2(\mathbf{y_2}-\mathbf{y_1})\nonumber\\-&\mathbf{x_1}+\mathbf{y_1})+\lambda_3-\lambda_4=0.
\end{align}
\par Considering KKT conditions, the solution when all the multipliers are 0 is obtained as follows:
\begin{align}
    \hat{k_1}&=\frac{\mathbf{u}\cdot\mathbf{v}\hspace{1mm}\mathbf{v}\cdot\mathbf{w}-||\mathbf{v}||_2^2\hspace{1mm}\mathbf{u}\cdot\mathbf{w}}{(\mathbf{u}\cdot\mathbf{v})^2-||\mathbf{u}||_2^2||\mathbf{v}||_2^2},\nonumber\\
    \hat{k_2}&=\frac{-\mathbf{u}\cdot\mathbf{v}\hspace{1mm}\mathbf{u}\cdot\mathbf{w}+||\mathbf{u}||_2^2\hspace{1mm}\mathbf{v}\cdot\mathbf{w}}{(\mathbf{u}\cdot\mathbf{v})^2-||\mathbf{u}||_2^2||\mathbf{v}||_2^2},
\end{align}
where $\mathbf{u}=\mathbf{x_1}-\mathbf{x_2}$, $\mathbf{v}=\mathbf{y_1}-\mathbf{y_2}$, and $\mathbf{w}=\mathbf{x_1}-\mathbf{y_1}$. The solutions to the remaining cases can be obtained by following a similar approach as presented in Section \ref{solution_section}.
\begin{figure}[h!]
    \begin{subfigure}[b]{0.32\columnwidth}
         \centering
         \includegraphics[scale=0.22]{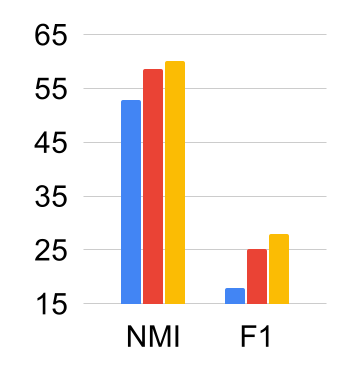}
    \caption{Clustering}
     \end{subfigure}
     \hfill
     \begin{subfigure}[b]{0.65\columnwidth}
         \centering
         \includegraphics[scale=0.2]{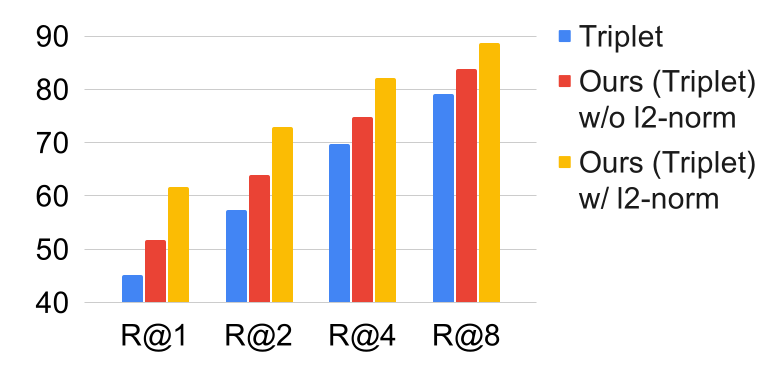}
    \caption{Retrieval}
     \end{subfigure}
\caption{Performance (\%) comparison of triplet, Ours (Triplet) without $l_2$-normalization and Ours (Triplet) with $l_2$-normalization for clustering and retrieval tasks. }
\label{l2norm}
\vspace{-2mm}
\end{figure}
\begin{table*}[h!]
    \centering
     \resizebox{2.05\columnwidth}{!}{\begin{tabular}{lccccccccccccccccc}
    \toprule
    &\multicolumn{6}{c}{CUB-200-2011} & \multicolumn{6}{c}{Cars196} & \multicolumn{5}{c}{SOP}\\
    \cmidrule(lr){2-7}\cmidrule(lr){8-13}\cmidrule(lr){14-18}
    Method & NMI & F$_1$& R@1 & R@2 & R@4 & R@8 &  NMI & F$_1$ & R@1 & R@2 & R@4 & R@8 &  NMI & F$_1$ & R@1 & R@10 & R@100 \\
    \midrule
      Triplet & 49.8 & 15.0 & 35.9 & 47.7 & 59.1 & 70.0 & 52.9 & 17.9 & 45.1 & 57.4 & 69.7 & 79.2 & 86.3 & 20.2 & 53.9 &72.1 & 85.7 \\
      
      DAML (Triplet) & 51.3 & 17.6 & 37.6 & 49.3 & 61.3 & 74.4 & 56.5 & 22.9 & 60.6 & 72.5 & \textbf{82.5} & \textbf{89.9} & 87.1 & 22.3 & 58.1 & 75.0 & 88.0\\
      HDML (Triplet) & 55.1 & 21.9 & 43.6 & 55.8 & 67.7 & 78.3 & 59.4 & 27.2 & 61.0 & 72.6 & 80.7 & 88.5 & 87.2 & 22.5 & 58.5 & 75.5 & 88.3 \\
      EE (Triplet) & 55.7 & 22.4 & 44.3 & 57.0 & 68.1 & 78.9 & \textbf{60.3} & 25.1 & 57.2 & 70.5 & 81.3 & 88.2 & 87.4 & 24.8 & 62.4 & 79.0 & 91.0 \\
      LoOp (Triplet) &  \textcolor{blue}{\textbf{59.9}} & \textcolor{blue}{\textbf{26.5}} & \textcolor{blue}{\textbf{50.3}} & \textcolor{blue}{\textbf{63.1}} & \textcolor{blue}{\textbf{74.4}} & \textcolor{blue}{\textbf{82.7}} &  \textcolor{blue}{60.2} & \textcolor{blue}{\textbf{28.0}} & \textcolor{blue}{\textbf{61.7}} & \textcolor{blue}{\textbf{72.9}} & \textcolor{blue}{82.1} & \textcolor{blue}{88.7} &  \textcolor{blue}{\textbf{88.7}} & \textcolor{blue}{\textbf{30.6}} & \textcolor{blue}{\textbf{69.6}} & \textcolor{blue}{\textbf{84.9}} & \textcolor{blue}{\textbf{93.7}}\\
    \midrule  
      HPHNtri & 58.1 & 24.2 & 48.3 & 61.9 & 73.0 & 82.3 & 57.4 & 22.6 & 60.3 & 73.4 & 83.5 & 90.5 & 91.4 & 43.3 & 75.5 & 88.8 & 95.4\\
      EE (HPHNtri)& 60.5 & 27.0 & 51.7 & 63.5 & 74.5 & 82.5 & 63.1 & 32.0 & 71.6 & 80.7 & 87.5 & 92.2 \ & \textbf{91.5} & \textbf{43.6} & 77.2 & 89.6 & 95.5\\
     LoOp (HPHNtri) & \textcolor{blue}{\textbf{61.2}} & \textcolor{blue}{\textbf{27.5}} & \textcolor{blue}{\textbf{51.8}} & \textcolor{blue}{\textbf{63.6}} & \textcolor{blue}{\textbf{74.9}} & \textcolor{blue}{\textbf{83.9}} & \textcolor{blue}{\textbf{64.8}} & \textcolor{blue}{\textbf{34.1}} & \textcolor{blue}{\textbf{74.5}} & \textcolor{blue}{\textbf{83.3}} & \textcolor{blue}{\textbf{89.4}} & \textcolor{blue}{\textbf{93.6}} 
     & 90.8 & 36.9 & \textcolor{blue}{\textbf{77.6}} & \textcolor{blue}{\textbf{90.2}} & \textcolor{blue}{\textbf{96.1}}\\
    \midrule
      LS & 56.4 & 22.6 & 46.9 & 59.8 & 71.2 & 81.5 & 57.8 & 25.1 & 59.9 & 70.4 & 79.6 & 87.0 & 87.2 & 25.3 & 62.6 & 80.9 & 91.2\\ 
      DAML (LS) & 59.5 & 26.6 & 49.0 & 62.2 & 73.7 & 83.3 & \textbf{63.1} & \textbf{31.9} & \textbf{72.5} & \textbf{82.1} & \textbf{88.5} & \textbf{92.9} & 89.1 & 31.7 & 66.3 & 82.8 & 92.5 \\
      EE (LS) & \textbf{61.2} & \textbf{28.2} & \textbf{54.2} & \textbf{66.6} & \textbf{76.7} & \textbf{85.2} & 59.1 & 27.2 & 65.2 & 76.4 & 85.6 & 89.5 & 89.6 & 35.3 & 70.6 & 85.5 & 93.6\\
      LoOp (LS) & \textcolor{blue}{60.3} & \textcolor{blue}{27.4} & \textcolor{blue}{52.8} & \textcolor{blue}{65.0} & \textcolor{blue}{76.2} & \textcolor{blue}{84.5} & 57.4 & \textcolor{blue}{25.9} & \textcolor{blue}{66.2} & \textcolor{blue}{76.8} & \textcolor{blue}{84.7} & \textcolor{blue}{89.8} & \textcolor{blue}{\textbf{89.8}} & \textcolor{blue}{\textbf{35.9}} & \textcolor{blue}{\textbf{77.1}} & \textcolor{blue}{\textbf{89.7}} & \textcolor{blue}{\textbf{95.6}}\\
    \midrule
      MS & 59.3 & 26.0 & 50.9 & 63.0 & 74.1 & 83.3 & \textbf{63.3} & \textbf{31.7} & 71.0 & 80.8 & 87.5 & 92.6 & 89.3 & 33.7 & 75.0 & 88.7 & 95.7\\
      LoOp (MS) & \textcolor{blue}{\textbf{61.1}} & \textcolor{blue}{\textbf{28.4}} & \textcolor{blue}{\textbf{52.0}} & \textcolor{blue}{\textbf{64.3}} & \textcolor{blue}{\textbf{75.0}} & \textcolor{blue}{\textbf{84.1}} &  63.0 & 30.6 & \textcolor{blue}{\textbf{72.6}} & \textcolor{blue}{\textbf{81.5}} & \textcolor{blue}{\textbf{88.4}} & \textcolor{blue}{\textbf{92.8}} & \textcolor{blue}{\textbf{89.4}} & \textcolor{blue}{\textbf{34.2}} & \textcolor{blue}{\textbf{76.6}} & \textcolor{blue}{\textbf{89.8}} & \textcolor{blue}{\textbf{95.8}}\\
    \bottomrule  
    \end{tabular}}
    \caption{Clustering and retrieval performance of generation-based methods for CUB-200-2011, Cars196 and SOP datasets. \textbf{Bold} numbers indicate the best values within each metric learning loss. \textcolor{blue}{Blue} numbers indicate cases where the combination of the proposed approach with a metric learning loss outperforms the baseline metric learning loss.}
    \label{tab:my_label}
    \vspace{-2mm}
\end{table*}
\begin{table}[h!]
     \begin{subtable}{\columnwidth}\centering
    {\resizebox{0.85\columnwidth}{!}{\begin{tabular}{lcccccc}
    \toprule
    Method & NMI & F$_1$& R@1 & R@2 & R@4 & R@8 \\
    \midrule
      Rand-disjoint & 49.8 & 15.0 & 35.9 & 47.7 & 59.1 & 70.0\\
      Semi-hard & 53.4 & 17.9 & 40.6 & 52.3 & 64.2 & 75.0\\
      DE-DSP & 53.7 & 19.8 & 41.0 & 53.2 & 64.8 & - \\
      Dis-weighted & 56.3 & 25.4 & 44.1 & 57.5 & 70.1 & 80.5\\
      Smart mining & 58.1 & - & 45.9 & 57.7 & 69.6 & 79.8\\
      LoOp &  \textcolor{blue}{\textbf{59.9}} & \textcolor{blue}{\textbf{26.5}} & \textcolor{blue}{\textbf{50.3}} & \textcolor{blue}{\textbf{63.1}} & \textcolor{blue}{\textbf{74.4}} & \textcolor{blue}{\textbf{82.7}} \\
    \bottomrule  
    \end{tabular}}}
    \vspace{-1mm}
    \subcaption{CUB-200-2011 dataset}
    \vspace{1mm}
    \end{subtable}
    \begin{subtable}{\columnwidth}\centering
    {\resizebox{0.85\columnwidth}{!}{\begin{tabular}{lcccccc}
    \toprule
    Method & NMI & F$_1$ & R@1 & R@2 & R@4 & R@8\\
    \midrule
      Rand-disjoint & 52.9 & 17.9 & 45.1 & 57.4 & 69.7 & 79.2 \\
     Semi-hard & 55.7 & 22.4 & 53.2 & 65.4 & 74.3 & 83.6  \\
     DE-DSP & 55.0 & 22.3 & 59.3 & 71.3 & 81.3 & -\\
     Dis-weighted & 58.3 & 25.4 & 59.4 & 72.3 & 81.6 & 87.2\\
     Smart mining & 58.2 & - & 56.1 & 68.3 & 78.0 & 85.9 \\
      LoOp & \textcolor{blue}{\textbf{60.2}} & \textcolor{blue}{\textbf{28.0}} & \textcolor{blue}{\textbf{61.7}} & \textcolor{blue}{\textbf{72.9}} & \textcolor{blue}{\textbf{82.1}} & \textcolor{blue}{\textbf{88.7}} \\
    \bottomrule  
    \end{tabular}}}
    \vspace{-1mm}
    \subcaption{Cars196 dataset}
    \vspace{1mm}
    \end{subtable}
    \begin{subtable}{\columnwidth}\centering
    {\resizebox{0.8\columnwidth}{!}{\begin{tabular}{lccccc}
    \toprule
    Method & NMI & F$_1$ & R@1 & R@10 & R@100\\
    \midrule
      Rand-disjoint & 86.3 & 20.2 & 53.9 &72.1 & 85.7 \\
     Semi-hard & 86.7 & 22.1 & 57.8 & 75.3 & 88.1 \\
     DE-DSP & 87.4 & 22.7 & 58.2 & 75.8 & 88.4\\
     Dis-weighted & 87.9 & 23.4 & 58.9 & 77.2 & 89.6 \\
      LoOp &  \textcolor{blue}{\textbf{88.7}} & \textcolor{blue}{\textbf{30.6}} & \textcolor{blue}{\textbf{69.6}} & \textcolor{blue}{\textbf{84.9}} & \textcolor{blue}{\textbf{93.7}} \\
    \bottomrule  
    \end{tabular}}}
    \vspace{-1mm}
    \subcaption{SOP dataset}
    \end{subtable}
    \caption{Comparison with clustering and retrieval performance of sampling-based methods for (a) CUB-200-2011, (b) Cars196 and (c) SOP datasets using triplet loss. \textbf{Bold} numbers indicate the best values. \textcolor{blue}{Blue} numbers indicate cases where the combination of LoOp and triplet loss outperforms the baseline. - indicates not reported.}
    \label{tab:my_label2}
    \vspace{-3mm}
\end{table}
\par The results of our approach for triplet loss with and without $l_2$-normalization as well as comparison with triplet loss (baseline) for Cars196 dataset are shown in Fig. \ref{l2norm}. It is observed that our approach performs better when the optimal distance is calculated between the curves rather than the line segments. This is because the points are $l_2$-normalized in the former case. Nevertheless, the latter case without $l_2$-normalization performs better than the baseline. This shows that the proposed approach to consider the minimum distance between two pairs of samples from different classes, greatly benefits the deep metric learning framework. 
\vspace{-1mm}
\subsection{Time and Memory Requirement}
\label{time}
\par Incorporating LoOp into metric learning losses leads to a minimal overhead in computation and time. For MS loss, the time per iteration changes from 0.071s to 0.121s. For HPHN-triplet loss, the time per iteration for LoOp is 0.058s. Thus, the time taken by LoOp remains in milliseconds for various metric learning losses.
\par In terms of computation and memory requirement, HPHN-triplet loss requires $O(B_S^2)$ combinations to compute the loss, which is same as the requirement of our approach. In contrast, the memory requirement of EE \cite{embex} depends on the number of interpolated points $n$. It needs to consider $O(n^2B_S^2)$ combinations to compute the loss. Thus, our method is both time and memory efficient. 
\vspace{-1mm}
\subsection{Comparison with State-of-the-art}
\label{comp_sota}
The results of our proposed approach and comparison with the baseline metric learning losses as well as SOTA methods for hard negative generation and sampling are shown in Tables \ref{tab:my_label} and \ref{tab:my_label2}, respectively. 
In Table \ref{tab:my_label}, it can be observed that for CUB-200-2011, LoOp outperforms the baseline metric learning losses in all cases. With the exception of EE for LS loss, our approach also shows better performance than the SOTA hard negative generation methods. The improvements with respect to previous best values range from 0.1\% to 6\% in Recall@1 and 0.7\% to 2.2\% in NMI. For Cars196 dataset, our approach provides a boost in performance for most of the baseline losses. The improvements with respect to previous best values range from 0.7\% to 2.9\% in Recall@1. For SOP dataset, it can be seen that for all cases, LoOp improves over the baseline losses. In terms of comparison with SOTA methods, there is an increase of upto 7.2\% in Recall@1, while NMI shows an increase of 0.1\% to 1.3\%. 
\par It is noteworthy that although HPHN-triplet and MS loss look for hard positive and negative samples as part of their formulation, LoOp further improves the results obtained by these methods. The maximum increase in Recall@1 value is 14.2\% for HPHN-triplet and 1.6\% for MS loss for the Cars196 dataset, as compared to the respective baselines. MS loss has a strict criterion that selects only the most informative samples, and hence the increase in performance is not very significant.  
\par Table \ref{tab:my_label2} presents the comparison with sampling methods, like random  sampling of disjoint tuples (rand-disjoint), semi-hard negative mining \cite{triplet1}, DE-DSP \cite{de_dsp}, distance-weighted sampling \cite{min1}, and smart mining \cite{min2}. 
It can be seen that LoOp outperforms all methods for all three datasets for both image clustering and retrieval tasks.
\vspace{-2mm}
\section{Conclusion}
\label{conc}
We propose an approach to find optimal hard negative embeddings for deep metric learning. This is done by calculating the points which minimize the distance between two curves, each representative of a different class. Our approach can be easily integrated with representative metric learning losses, avoids the computational cost of generating hard data samples, and utilizes the training set effectively unlike hard negative mining-based methods. We demonstrated that our approach boosts the performance of existing metric learning losses and outperforms state-of-the-art hard negative sampling and generation-based methods.

{\small
\bibliographystyle{ieee_fullname}
\bibliography{egbib}
}

\pagebreak
\setcounter{equation}{0}
\setcounter{figure}{0}
\setcounter{table}{0}
\setcounter{page}{1}
\setcounter{section}{0}

\twocolumn[{%
 \centering
 \large \textbf{Supplementary Material for\\ LoOp: Looking for Optimal Hard Negative \\ Embeddings for Deep Metric Learning}
 \vspace{5mm}
}]

\begin{figure*}[!th]
    \centering
    \begin{subfigure}[b]{0.66\columnwidth}
    \centering
    \includegraphics[scale=0.26]{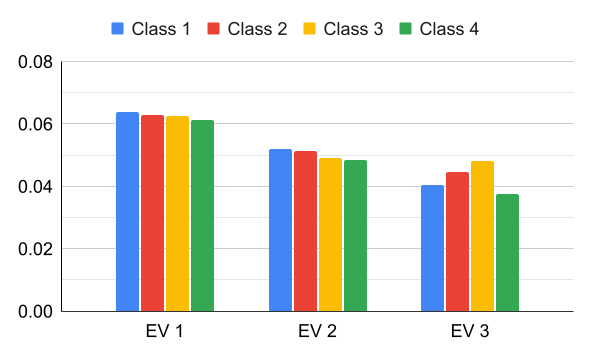}
    \caption{CUB-200-2011 dataset.}
    \end{subfigure}
    \hfill
    \begin{subfigure}[b]{0.66\columnwidth}
    \centering
    \includegraphics[scale=0.26]{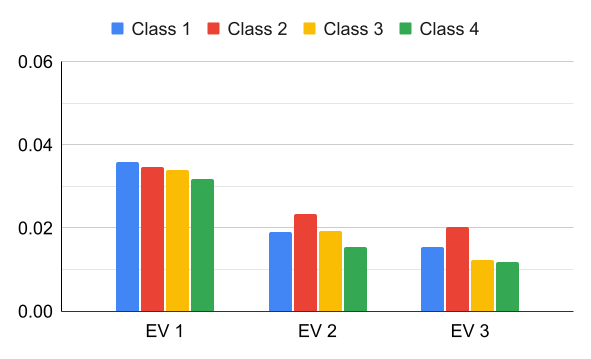}
    \caption{Cars196 dataset.}
    \end{subfigure}
    \hfill
    \begin{subfigure}[b]{0.66\columnwidth}
    \centering
    \includegraphics[scale=0.26]{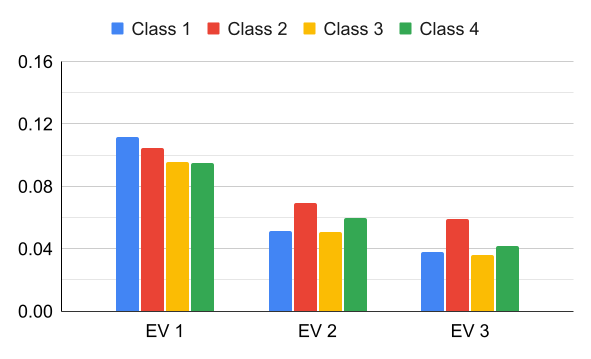}
    \caption{SOP dataset.}
    \end{subfigure}
    \caption{Representation of eigenvalues of the covariance matrices of four randomly picked classes for (a) CUB-200-2011, (b) Cars196, and (c) SOP datasets. The embeddings have been projected to a 3D space using PCA. The model is trained using the proposed approach with triplet loss. The values from the same position on the diagonal have been plotted together.}
    \label{val_sph}
\end{figure*}

\vspace{5mm}
\section{Introduction}
\vspace{-1mm}
This supplementary material includes the results for validation of the spherical-homoscedasticity assumption (Section \ref{validation}), proof of Proposition 1 (Section \ref{prop1s}), the proofs for obtaining the optimal hard negatives (Section \ref{proofs}), the toy examples demonstrating the effectiveness of using optimal hard negatives (Section \ref{toy}), the t-SNE plots for visualizing the training process (Section \ref{tsne}), some results demonstrating the effect of network architecture (Section \ref{more_res}), and results for train-validate-test split (Section \ref{tvt}).
\vspace{-1mm}
\section{Validation of Spherical-homoscedastic Distributions Assumption}
\label{validation}
\vspace{-1mm}
\par To validate the spherical-homoscedasticity assumption, we project the embeddings (of samples belonging to the same class) to a 3D space using principal component analysis (PCA). We need to show that the eigenvalues of the covariance matrices of the data distributions of different classes are close to one another. Fig. \ref{val_sph} shows the eigenvalues for four randomly picked classes. It is seen that these eigenvalues are very close indeed for all the three datasets, indicating a similar shape of data distributions (spherical-homoscedasticity). 
\par Further, we calculate the mean and standard deviation ($\times$100) of the three eigenvalues obtained by using all the classes in each dataset. They are 7.02$\pm$1.84, 4.84$\pm$1.05, 3.84$\pm$0.84 for the CUB-200-2011 dataset, 3.67$\pm$1.17, 2.16$\pm$0.60, 1.56$\pm$0.41 for the Cars196 dataset, and 12.24$\pm5.25$, 6.63$\pm2.98$, 3.93$\pm2.04$ for the SOP dataset. The values of the standard deviations are small, which again validates the assumption.
\vspace{-1mm}
\section{Proof of Proposition 1}
\label{prop1s}
\vspace{-1mm}
\begin{customprop}{1}\label{prop1}
If the pairs of points $\mathbf{x_1}$, $\mathbf{x_2}$ and $\mathbf{y_1}$, $\mathbf{y_2}$ (from two different classes) belong to spherical-homoscedastic distributions, then the points on the curves $\arc{\mathbf{x_1x_2}}$ and $\arc{\mathbf{y_1y_2}}$ have a higher probability of belonging to the same classes as $\mathbf{x_1}$, $\mathbf{x_2}$ and $\mathbf{y_1}$, $\mathbf{y_2}$, respectively, than the other classes.
\end{customprop}
\vspace{-2mm}
\begin{proof}
 The definition of spherical-homoscedastic distributions says that such distributions are separated by decision boundaries that are hyperplanes. In such a case, a region between two points (say $\mathbf{x_1}$, $\mathbf{x_2}$) of the same class will also belong to that class with a higher probability than the other classes. Thus, any points sampled from that region will have the highest probability of belonging to the same class as that of $\mathbf{x_1}$, $\mathbf{x_2}$. As we assign class belongingness of an unknown point based on the maximum probability or likelihood, the points lying on the curves $\arc{\mathbf{x_1x_2}}$ and $\arc{\mathbf{y_1y_2}}$ will belong to the same classes as $\mathbf{x_1}$, $\mathbf{x_2}$ and $\mathbf{y_1}$, $\mathbf{y_2}$, respectively.
\end{proof}
\vspace{-2mm}
\section{Proofs for Finding Optimal Hard Negatives}
\label{proofs}
\vspace{-1mm}
In this section, we present the proofs for finding the optimal points $\mathbf{p_1}$ and $\mathbf{p_2}$, such that the distance between a pair of positives and a pair of negatives is minimized. Section \ref{l2yo} presents the case where the optimal points are $l_2$-normalized and lie on the hypersphere, whereas Section \ref{l2no} presents the case where optimal points are not $l_2$-normalized. \vspace{-1mm}
\subsection{Proof for solution of optimal points with $l_2$-normalization}
\label{l2yo}
\vspace{-1mm}
First, the expression for $\mathbf{p_1}$ is obtained as follows:
\vspace{-1mm}
\begin{align*}
    \mathbf{p_1}&=\mathbf{Rx_1}=\mathbf{Rn_1}\\
    &=(\mathbf{I}+\sin\alpha(\mathbf{n_2}\mathbf{n_1}^T-\mathbf{n_1}\mathbf{n_2}^T)\\
    &-(1-\cos\alpha)(\mathbf{n_1}\mathbf{n_1}^T+\mathbf{n_2}\mathbf{n_2}^T))\mathbf{n_1}\\
    &=\mathbf{n_1}+(\mathbf{n_2}\cancelto{1}{\mathbf{n_1^Tn_1}}-\mathbf{n_1}\cancelto{0}{\mathbf{n_2^Tn_1}})\sin\alpha\\
    &-(\mathbf{n_1}\cancelto{1}{\mathbf{n_1^Tn_1}}+\mathbf{n_2}\cancelto{0}{\mathbf{n_2^Tn_1}})(1-\cos\alpha)\\
    &=\mathbf{n_1}+\mathbf{n_2}\sin\alpha-\mathbf{n_1}(1-\cos\alpha)\\
    &=\mathbf{n_1}\cos\alpha+\mathbf{n_2}\sin\alpha,
\end{align*}
where the dot products are reduced to 1 or 0 as $\mathbf{n_1}$ and $\mathbf{n_2}$ are orthonormal vectors. 

The KKT conditions to obtain the solution for the problem listed in Section 2.2 of the paper are mentioned again for reference:
\begin{align}
\label{delLdelalpha}
\frac{\partial L}{\partial \alpha}&=\frac{\partial f}{\partial \alpha} + \lambda_1 - \lambda_2 \\
\nonumber&=a \cos\alpha\sin\beta -b\sin\alpha\sin\beta+c\cos\alpha\cos\beta\\
\label{delLdelalpha1}
&-d\sin\alpha\cos\beta+\lambda_1-\lambda_2=0,\\
\label{delLdelbeta}
\frac{\partial L}{\partial \beta}&=\frac{\partial f}{\partial \beta} + \lambda_3 - \lambda_4\\ 
&=a \sin\alpha\cos\beta+b\cos\alpha\cos\beta-c\sin\alpha\sin\beta\nonumber\\
\label{delLdelbeta1}
&-d\cos\alpha\sin\beta+\lambda_3-\lambda_4=0,
\end{align}
\vspace{-5mm}
\begin{align}
\label{comp_slack}
\lambda_ig_i&=0 \text{ ; } i=1,2,3,4,\\
\label{lambdas0}
\lambda_i&\leq 0 \text{ ; } i=1,2,3,4,\\
\label{gs0}
g_i&\leq 0 \text{ ; } i=1,2,3,4.
\end{align}
The solutions for the 9 possible cases to be considered are discussed below. 

\textbf{Note}: The solutions to the equations involving $\alpha, \beta$ are denoted by $\hat{\alpha}, \hat{\beta}$.

\textbf{Case 0}: $\lambda_1=0, \lambda_2=0, \lambda_3=0, \lambda_4=0$.

Using (\ref{delLdelalpha1}), (\ref{delLdelbeta1}), we obtain:
\begin{align}
    a \cos\alpha\sin\beta -b\sin\alpha\sin\beta+c\cos\alpha\cos\beta\nonumber\\
    \label{case00}-d\sin\alpha\cos\beta =0, \\
    a \sin\alpha\cos\beta+b\cos\alpha\cos\beta-c\sin\alpha\sin\beta\nonumber\\
    \label{case0a}-d\cos\alpha\sin\beta=0.
    \end{align}
Simplifying these, we get:
\begin{align}
    \label{case0b}\cos\alpha(a\sin\beta+c\cos\beta) = \sin\alpha(b\sin\beta+d\cos\beta),\\
     \label{case0c}\cos\alpha(b\cos\beta-d\sin\beta) = \sin\alpha(c\sin\beta-a\cos\beta).
\end{align}
Dividing both sides in (\ref{case0b}), (\ref{case0c}) by $\cos\alpha \cos\beta$, and then dividing (\ref{case0b}) by (\ref{case0c}) we get:
\begin{align*}
    \frac{a\tan\beta+c}{b-d\tan\beta} = \frac{b\tan\beta+d}{c\tan\beta-a}.
    \end{align*}
This can be simplified to get:
\begin{align*}
     \tan^2\beta - \frac{a^2+b^2-c^2-d^2}{ac+bd}\tan\beta -1 =0,\\
    \therefore \hat\beta=\tan^{-1}\left(\frac{B\mp\sqrt{B^2+4(ac+bd)^2}}{2(ac+bd)}\right).
\end{align*}
For $\hat{\beta}<0$, we consider $\hat{\beta} + \pi$ as a possible solution due to the constraint $g_3$.

Similarly, using (\ref{case00}), (\ref{case0a}), we can form a quadratic equation in $\tan\alpha$:
\begin{align*}
     \tan^2\alpha - \frac{a^2-b^2-c^2+d^2}{ab+cd}\tan\alpha -1 =0, \\
      \therefore \hat\alpha=\tan^{-1}\left(\frac{A\pm\sqrt{A^2+4(ab+cd)^2}}{2(ab+cd)}\right).
\end{align*}
Similar to the argument for $\hat{\beta}$, $\hat{\alpha}+\pi$ is considered as a possible solution when $\hat{\alpha}<0$. (In the aforementioned equations, $A=a^2-b^2+c^2-d^2$ and $B=a^2+b^2-c^2-d^2$.)

\textbf{Case 1}: $g_1=0$ $(\hat{\alpha}=0), \lambda_2=0, \lambda_3=0, \lambda_4=0$.
\begin{align}
    \nonumber\frac{\partial L}{\partial \beta} &= b\cos\beta -d\sin\beta = 0, \implies \tan\beta = \frac{b}{d}, \\ 
    \label{case11} \therefore \hat{\beta} &= \tan^{-1}\left(\frac{b}{d}\right).
\end{align}
Using (\ref{delLdelalpha}), (\ref{delLdelalpha1}), (\ref{case11}), we get: 
\begin{align*}
    \lambda_1 = -\left.\frac{\partial f}{\partial \alpha}\right|_{\alpha=0, \beta=\hat\beta}
\end{align*}

\textbf{Case 2}: $g_2=0$ $(\hat{\alpha}=\alpha_0), \lambda_1=0, \lambda_3=0, \lambda_4=0$.

Using (\ref{delLdelbeta1}), we obtain:
\begin{align}
   \nonumber \frac{\partial L}{\partial \beta} &= a \sin\alpha_0\cos\beta  +  b\cos\alpha_0\cos\beta- 
     c\sin\alpha_0\sin\beta \\ 
     \nonumber&-d\cos\alpha_0\sin\beta=0, \\ 
    \nonumber&\implies  \tan \beta = \left(\frac{a\sin\alpha_0+b\cos\alpha_0}{c\sin\alpha_0+d\cos\alpha_0}\right), \\
   \label{case22}
   &\therefore \hat{\beta} = \tan^{-1}\left(\frac{a\sin\alpha_0+b\cos\alpha_0}{c\sin\alpha_0+d\cos\alpha_0}\right). 
\end{align}

Using (\ref{delLdelalpha}), (\ref{delLdelalpha1}), (\ref{case22}), we get:
\begin{align*}
    \lambda_2 = \left.\frac{\partial f}{\partial \alpha}\right|_{\alpha=\alpha_0, \beta=\hat\beta}
\end{align*}

\textbf{Case 3}: $g_3=0$ $(\hat{\beta}=0), \lambda_1=0, \lambda_2=0, \lambda_4=0$.

Using (\ref{delLdelalpha1}), we obtain:
\begin{align}
    \nonumber &c\cos\alpha -d\sin\alpha = 0, \implies \tan \alpha = \left(\frac{c}{d}\right), \\
    \label{case33} &\therefore \hat{\alpha} = \tan^{-1} \left(\frac{c}{d}\right).
\end{align}

Using (\ref{delLdelbeta}), (\ref{delLdelbeta1}), (\ref{case33}), we get:
\begin{align*}
    \lambda_3 = -\left.\frac{\partial f}{\partial \beta}\right|_{\alpha=\hat\alpha,\beta=0}
\end{align*}
\textbf{Case 4}: $g_4=0$ $(\hat{\beta}=\beta_0), \lambda_1=0, \lambda_2=0, \lambda_3=0$.

Using (\ref{delLdelalpha1}), we obtain:
\begin{align}
    \nonumber&a \cos\alpha\sin\beta_0 -b\sin\alpha\sin\beta_0+c\cos\alpha\cos\beta_0\\
    \nonumber&-d\sin\alpha\cos\beta_0 = 0, \\
    \nonumber& \implies \tan\alpha =\frac{a\sin\beta_0+c\cos\beta_0}{b\sin\beta_0+d\cos\beta_0}, \\
\label{case44}
&\therefore \hat{\alpha} = \tan^{-1}\left(\frac{a\sin\beta_0+c\cos\beta_0}{b\sin\beta_0+d\cos\beta_0}\right).
\end{align}

Using (\ref{delLdelbeta}), (\ref{delLdelbeta1}), (\ref{case44}), we obtain:
\begin{align*}
    \lambda_4 = \left.\frac{\partial f}{\partial \beta}\right|_{\alpha=\hat\alpha,\beta=\beta_0}
\end{align*}

\textbf{Case 5}: $g_1=0$ $(\hat{\alpha} =0),$ $g_3=0$ $(\hat{\beta}=0), \lambda_2=0, \lambda_4=0$.

Using (\ref{delLdelalpha}), (\ref{delLdelalpha1}), we have:
\begin{align*}
    \nonumber\frac{\partial L}{\partial \alpha}&=\frac{\partial f}{\partial \alpha} + \lambda_1=0 \\
    & \lambda_1 = -\left.\frac{\partial f}{\partial \alpha}\right|_{\alpha=0,\beta=0} 
\end{align*}

Similarly using (\ref{delLdelbeta}), (\ref{delLdelbeta1}), we get:

\begin{align*}
    \nonumber\frac{\partial L}{\partial \beta}&=\frac{\partial f}{\partial \beta} + \lambda_3=0 \\
    & \lambda_3 = -\left.\frac{\partial f}{\partial \beta}\right|_{\alpha=0,\beta=0} 
\end{align*}

For cases 6-8, we can easily obtain the respective $\lambda_i$ values in a similar fashion to case 5. These have already been listed in Table 1 in the paper. For the aforementioned cases, we consider $\hat{\alpha} (\hat{\beta})+\pi$ as a possible solution when $\hat{\alpha} (\hat{\beta})<0$ due to the constraint $g_1 (g_3)$.

\subsection{Proof for solution of optimal points without $l_2$-normalization}
\label{l2no}
In this case, $f$ is obtained as:
\begin{align*}
    f&(k_1,k_2)=||\mathbf{p_1}-\mathbf{p_2}||_2^2\\
    &=||k_1(\mathbf{x_2}-\mathbf{x_1})-k_2(\mathbf{y_2}-\mathbf{y_1})+\mathbf{x_1}-\mathbf{y_1}||_2^2\\
    &=||k_1\mathbf{u}-k_2\mathbf{v}-\mathbf{w}||_2^2.
\end{align*}
The Lagrangian function is given by:
\begin{equation*}
    L(k_1,k_2,\lambda_1,\lambda_2,\lambda_3,\lambda_4)=f(k_1,k_2)-\sum_{i=1}^{4}\lambda_ig_i.
\end{equation*}
The constraints are mentioned below for reference:
\begin{align}
\label{giforl2no}
    g_1&=-k_1\leq0 \text{ ; } g_2=k_1-1\leq0,\nonumber\\
    g_3&=-k_2\leq0 \text{ ; } g_4=k_2-1\leq0.
\end{align}
The partial derivatives for the KKT conditions are given as follows:
\begin{align}
   \label{deldelk1}
    \frac{\partial L}{\partial k_1}=&\frac{\partial f}{\partial k_1} + \lambda_1 - \lambda_2\\
    \label{deldelk11}
    =&\mathbf{u}\cdot(k_1\mathbf{u}-k_2\mathbf{v}-\mathbf{w})+\lambda_1-\lambda_2=0,\\
    \label{deldelk2}
   \frac{\partial L}{\partial k_2}=&\frac{\partial f}{\partial k_2} + \lambda_3 - \lambda_4\\
   \label{deldelk21}
   =&\mathbf{v}\cdot(-k_1\mathbf{u}+k_2\mathbf{v}+\mathbf{w})+\lambda_3-\lambda_4=0.
\end{align}
The rest of the KKT conditions are given by (\ref{comp_slack}), (\ref{lambdas0}), and (\ref{gs0}), where $g_i$'s are the ones mentioned in (\ref{giforl2no}). In a simplified form, (\ref{deldelk11}) and (\ref{deldelk21}) can be written as:
\begin{align}
    \label{deldelk1a} 
    \frac{\partial L}{\partial k_1}
    &= ak_1 + bk_2 + c + \lambda_1 - \lambda_2 =0,\\
    \label{deldelk2a}
    \frac{\partial L}{\partial k_2}
    &=a^{\prime}k_1 + b^{\prime}k_2 + c ^{\prime} + \lambda_3 - \lambda_4 = 0,
\end{align}
respectively, where $a=\mathbf{u}\cdot\mathbf{u}$, $b=-\mathbf{u}\cdot\mathbf{v}$, $c=-\mathbf{u}\cdot\mathbf{w}$, $a'=-\mathbf{v}\cdot\mathbf{u}$, $b'=\mathbf{v}\cdot\mathbf{v}$, $c'=\mathbf{v}\cdot\mathbf{w}$. The solutions for the 9 possible cases are discussed below.

\textbf{Note}: The solutions to the equations involving $k_1, k_2$ are denoted by $\hat{k_1}, \hat{k_2}$.

\textbf{Case 0}: $\lambda_1=0, \lambda_2=0, \lambda_3=0, \lambda_4=0$.

Using (\ref{deldelk1a}), (\ref{deldelk2a}) we get:
\begin{align*}
    &ak_1 + bk_2 + c =0, \\
    &a^{\prime}k_1 + b^{\prime}k_2 + c ^{\prime} = 0.
\end{align*}

Solving for $k_1, k_2$ we get:
\begin{align*}
    \hat{k_1} = \frac{b^{\prime}c- bc^{\prime}}{a^{\prime}b-ab^{\prime}}, \hspace{2mm} \hat{k_2} = \frac{ac^{\prime}- a^{\prime}c}{a^{\prime}b-ab^{\prime}}.
\end{align*}

\textbf{Case 1}: $g_1 =0$ $(\hat{k_1}=0), \lambda_2=0, \lambda_3=0, \lambda_4=0$.

Using (\ref{deldelk2a}), we get:
\begin{align}
    \label{case1} b^{\prime}k_2 &+ c^{\prime}  = 0, \nonumber\\
   \therefore \hat{k_2} &= \frac{-c^{\prime}}{b^{\prime}}.
\end{align}

Using (\ref{deldelk1a}), (\ref{deldelk1}), (\ref{case1}), we get:
\begin{align*}
    \lambda_1 = -\left.\frac{\partial f}{\partial k_1 }\right|_{k_1=0,k_2=\hat{k_2}}
\end{align*}

\textbf{Case 2}: $g_2 =0$ $(\hat{k_1}=1), \lambda_2=0, \lambda_3=0, \lambda_4=0$.

Using (\ref{deldelk2a}), we get:
\begin{align}
   \label{case2} a^{\prime} + b^{\prime}k_2 + c ^{\prime} = 0, \nonumber\\
   \therefore \hat{k_2} = -\frac{a^{\prime}+c^{\prime}}{b^{\prime}}.
\end{align}
Using (\ref{deldelk1a}), (\ref{deldelk1}), (\ref{case2}), we obtain:
\begin{align*}
    \lambda_2 = \left.\frac{\partial f}{\partial k_1 }\right|_{k_1=1,k_2=\hat{k_2}}
\end{align*}

Similar to the aforementioned 2 cases we can obtain the following results for case 3, and case 4. \\

\textbf{Case 3}: $g_3 =0$ $(\hat{k_2}=0), \lambda_1=0, \lambda_2=0, \lambda_4=0$.

Using (\ref{deldelk1a}), (\ref{deldelk2a}), (\ref{deldelk2}), we get:
\begin{align*}
    \hat{k_1}  = -\frac{c}{a} \implies \lambda_3 = - \left.\frac{\partial f}{\partial k_2 }\right|_{k_1=\hat{k_1},k_2=0}
\end{align*}

\textbf{Case 4}: $g_4 =0$ $(\hat{k_2}=1), \lambda_1=0, \lambda_2=0, \lambda_3=0$.

Using (\ref{deldelk1a}), (\ref{deldelk2a}), (\ref{deldelk2}) we get:
\begin{align*}
    \hat{k_1}  = -\frac{b+c}{a} \implies \lambda_3 = \left.\frac{\partial f}{\partial k_2 }\right|_{k_1=\hat{k_1},k_2=1}
\end{align*}

\textbf{Case 5}: $g_1 =0$ $(\hat{k_1}=0), g_3 =0$ $(\hat{k_2}=0), \lambda_2=0, \lambda_4=0$.

Using (\ref{deldelk1a}), (\ref{deldelk1}), and (\ref{deldelk2a}), (\ref{deldelk2}), we get:
\begin{align*}
    \lambda_1 = -\left.\frac{\partial f}{\partial k_1}\right|_{k_1=0,k_2=0} \\
    \lambda_3 =  -\left.\frac{\partial f}{\partial k_2}\right|_{k_1=0,k_2=0}
\end{align*}

\textbf{Case 6}: $g_1 =0$ $(\hat{k_1}=0), g_4 =0$ $(\hat{k_2}=1), \lambda_2=0, \lambda_3=0$.

Using (\ref{deldelk1a}), (\ref{deldelk1}), and (\ref{deldelk2a}), (\ref{deldelk2}), we get:
\begin{align*}
    \lambda_1 = -\left.\frac{\partial f}{\partial k_1}\right|_{k_1=0,k_2=1} \\
    \lambda_4 =  \left.\frac{\partial f}{\partial k_2}\right|_{k_1=0,k_2=1}
\end{align*}

\textbf{Case 7}: $g_2 = 0$ $(\hat{k_1}=1), g_3 =0$ $(\hat{k_2}=0), \lambda_1=0, \lambda_4=0$.

Using (\ref{deldelk1a}), (\ref{deldelk1}), and (\ref{deldelk2a}), (\ref{deldelk2}), we get:
\begin{align*}
    \lambda_2 = \left.\frac{\partial f}{\partial k_1}\right|_{k_1=1,k_2=0} \\
    \lambda_3 =  -\left.\frac{\partial f}{\partial k_2}\right|_{k_1=1,k_2=0}
\end{align*}

\textbf{Case 8}: $g_2=0$ $(\hat{k_1}=1), g_4=0$ $(\hat{k_2}=1), \lambda_1=0, \lambda_3=0$.

Using (\ref{deldelk1a}), (\ref{deldelk1}), and (\ref{deldelk2a}), (\ref{deldelk2}), we get:
\begin{align*}
    \lambda_2 = \left.\frac{\partial f}{\partial k_1}\right|_{k_1=1,k_2=1} \\
    \lambda_4 =  \left.\frac{\partial f}{\partial k_2}\right|_{k_1=1,k_2=1}
\end{align*}
\section{Analysis of Triplet Loss with Optimal Hard Negatives}
\label{toy}
Given two pairs of points in the embedding space: $\mathbf{x_1}$, $\mathbf{x_2}$ and $\mathbf{y_1}$, $\mathbf{y_2}$, belonging to two different classes. The triplet loss is formulated as follows:
\begin{align}
\mathcal{L}_{Tri}=\left[d(\mathbf{x_1},\mathbf{x_2})-d(\mathbf{x_1},\mathbf{y_1})+m\right]_+.
\end{align}

\begin{figure*}[h!]
    \centering
    \includegraphics[scale=0.45]{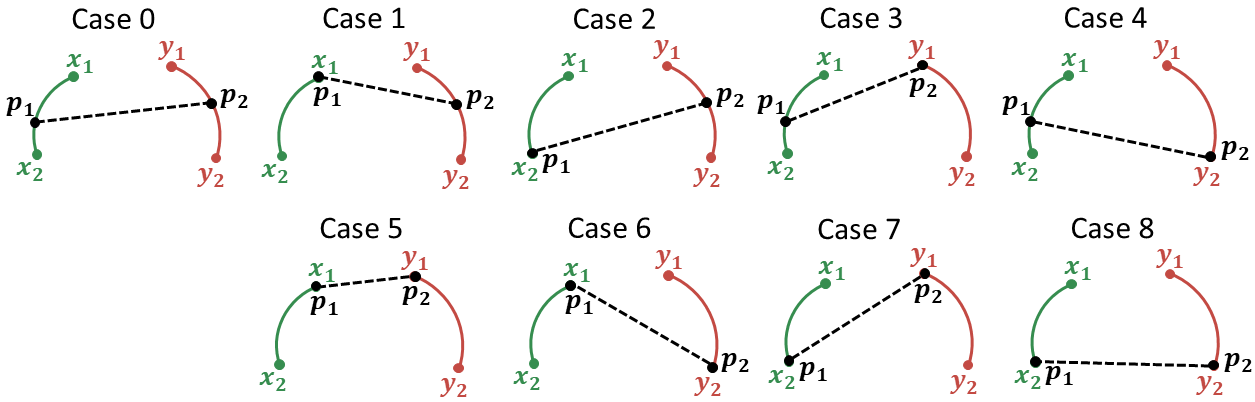}
    \caption{Illustrations of the 9 cases in which the KKT conditions for finding the minimum distance between arcs $\arc{\mathbf{x_1x_2}}$ and $\arc{\mathbf{y_1y_2}}$ can be satisfied.}
    \label{cases_fig}
\end{figure*}
\begin{figure*}[h!]
    \centering
    \includegraphics[scale=0.67]{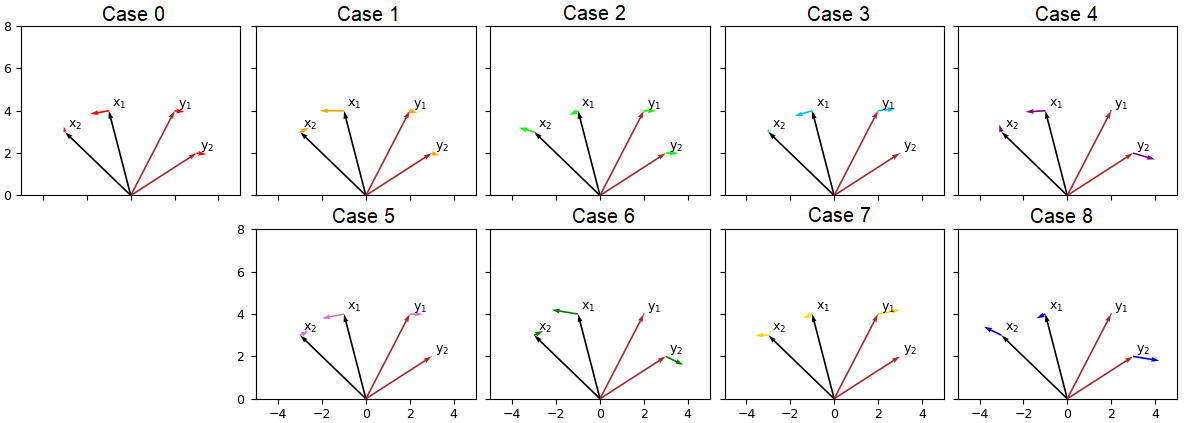}
    \caption{An example showing four samples and their updated positions obtained by using the gradients for different choices of $\mathbf{p_1}$ and $\mathbf{p_2}$. The samples have been selected such that $\mathbf{x_1}$ and $\mathbf{y_1}$ are the hardest negatives.}
    \label{ex1}
\end{figure*}
\begin{figure*}[h!]
    \centering
    \includegraphics[scale=0.67]{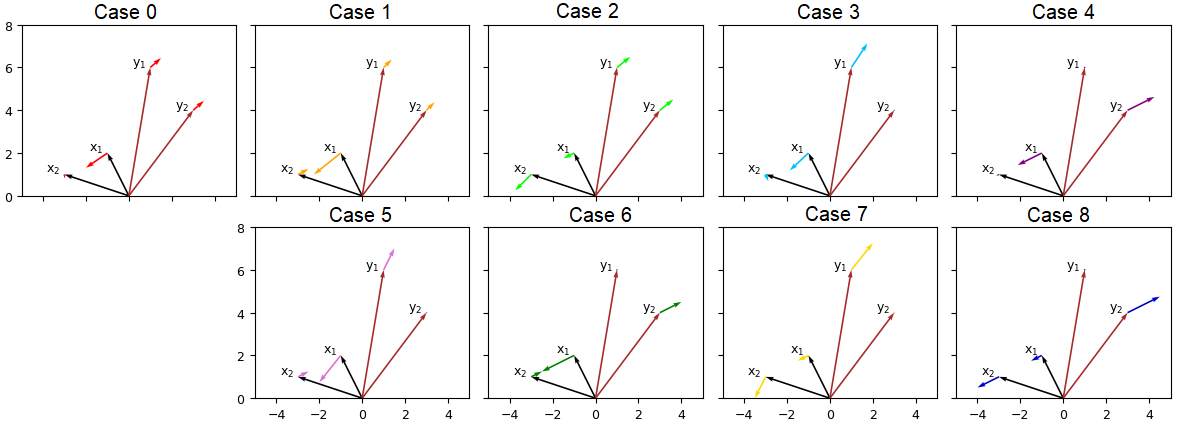}
    \caption{An example showing four samples and their updated positions obtained by using the gradients for different choices of $\mathbf{p_1}$ and $\mathbf{p_2}$. The samples have been selected such that $\mathbf{x_1}$ and $\frac{\mathbf{y_1}+\mathbf{y_2}}{2}$ are the hardest negatives.}
    \label{ex2}
\end{figure*}
\begin{figure*}[t]
    \centering
    \begin{subfigure}[b]{0.69\columnwidth}
         \centering
         \includegraphics[scale=0.31]{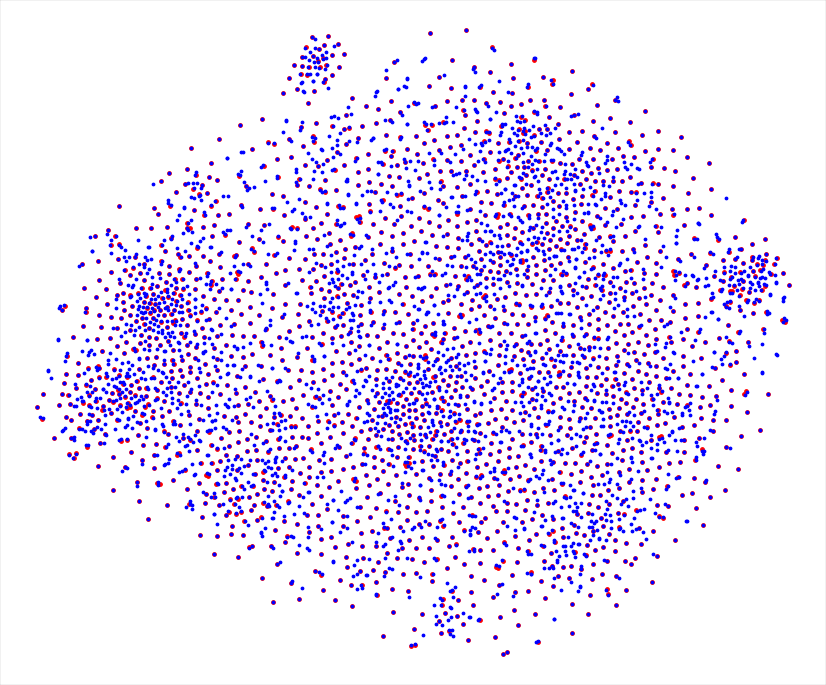}
    \caption{Epoch 10.}
     \end{subfigure}
     \centering
     \begin{subfigure}[b]{0.69\columnwidth}
         \centering
         \includegraphics[scale=0.31]{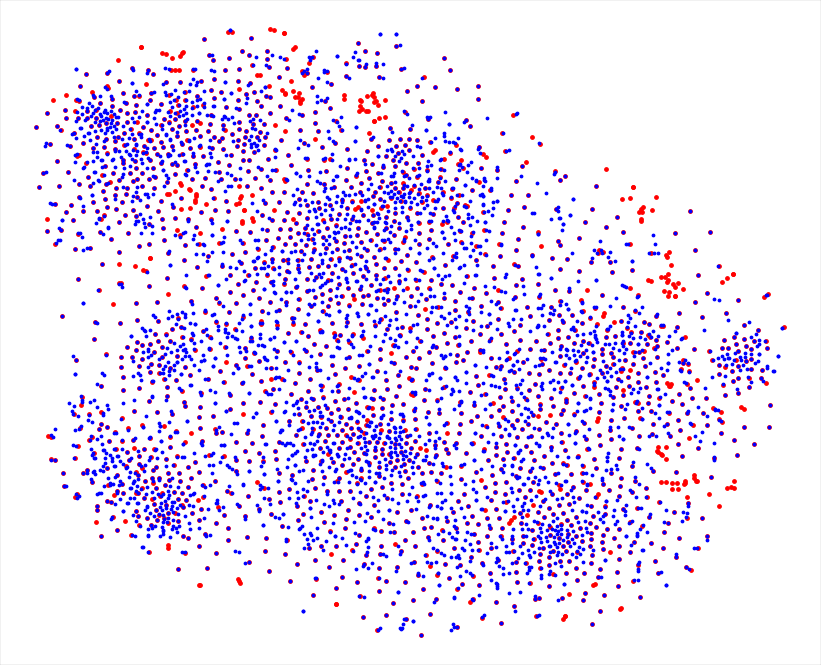}
    \caption{Epoch 50.}
     \end{subfigure}
     \centering
    \begin{subfigure}[b]{0.69\columnwidth}
         \centering
         \includegraphics[scale=0.31]{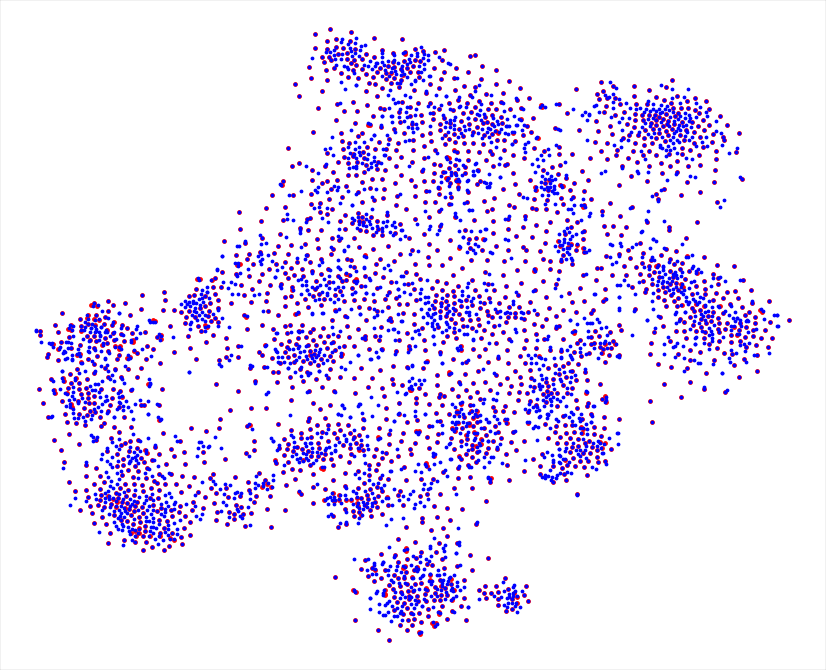}
    \caption{Epoch $\sim$ 200.}
     \end{subfigure}\\
     \centering
    \begin{subfigure}[b]{0.69\columnwidth}
         \centering
         \includegraphics[scale=0.31]{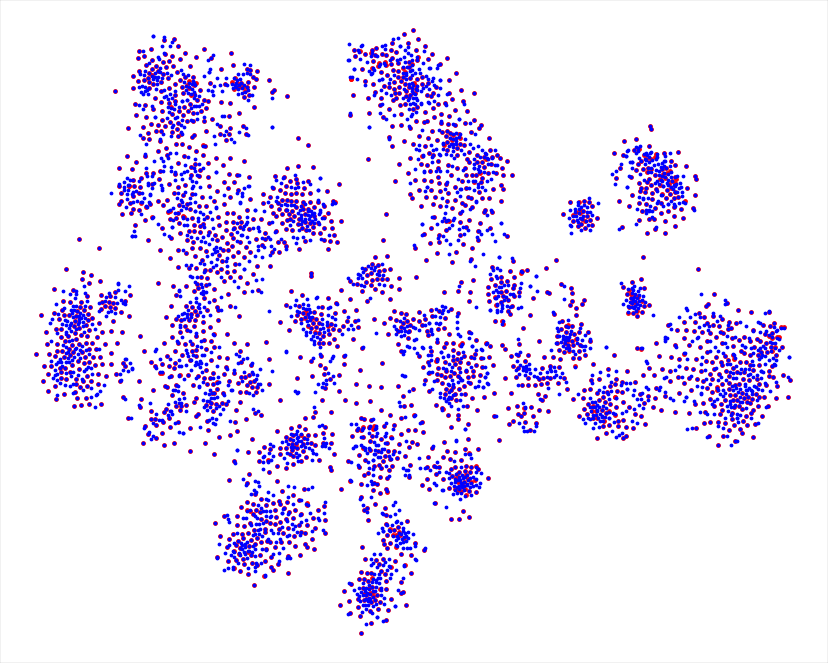}
    \caption{Epoch $\sim$ 400.}
     \end{subfigure}
     \centering
     \begin{subfigure}[b]{0.69\columnwidth}
         \centering
         \includegraphics[scale=0.31]{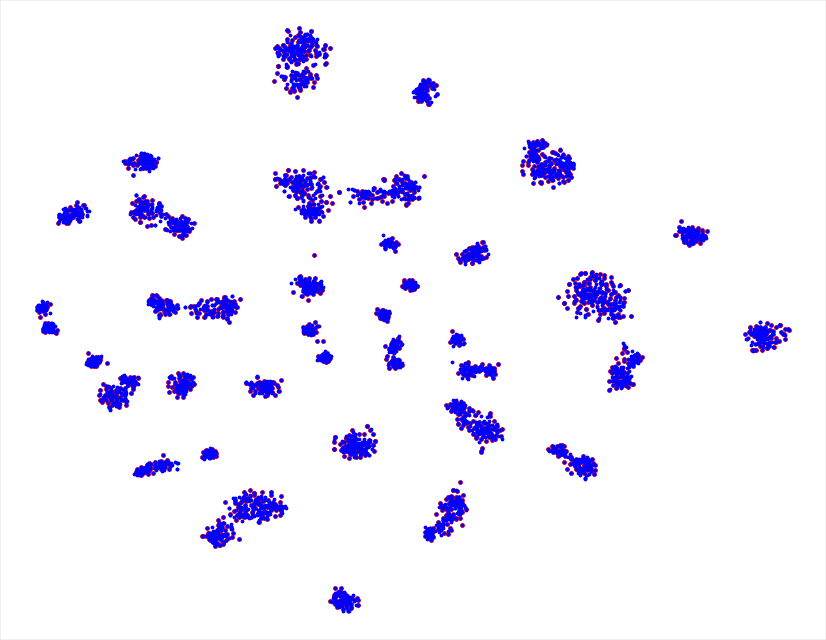}
    \caption{Epoch $\sim$ 4000.}
     \end{subfigure}
     \centering
     \begin{subfigure}[b]{0.69\columnwidth}
         \centering
         \includegraphics[scale=0.31]{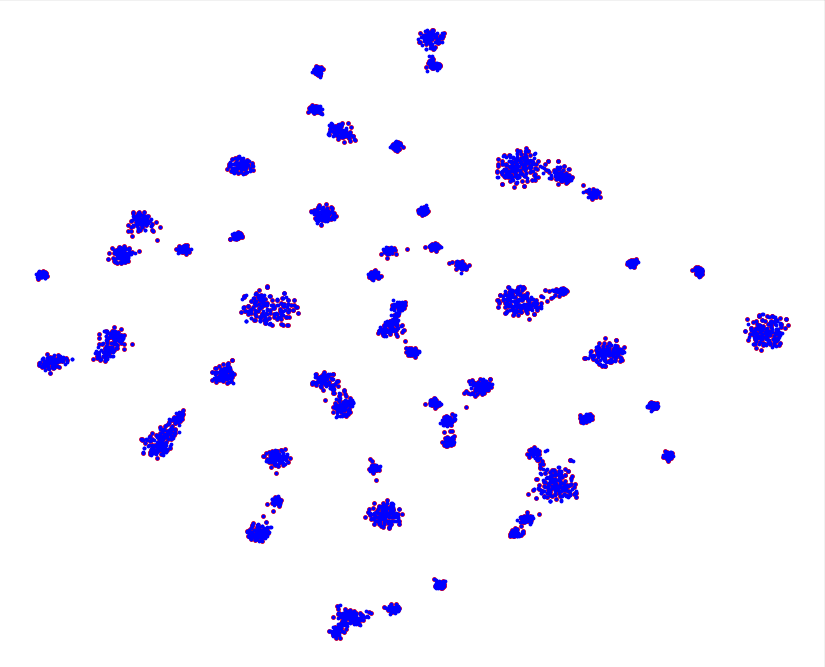}
    \caption{Epoch $\sim$ 8500.}
     \end{subfigure}
\caption{t-SNE visualization of LoOp with triplet loss using CARS196 dataset showing the embeddings of the train data for different epochs. \textcolor{blue}{Blue} samples are the original training samples, while \textcolor{red}{Red} samples are the synthetic samples generated by LoOp. The marker size of synthetic samples is bigger to improve visibility. As we move from (a) to (f), the red samples become less and less visible as they lie in the same region as the original samples. }
\label{tsne_plots}
\end{figure*}

The modified triplet loss obtained by incorporating the proposed approach is given by:
\begin{align}
\mathcal{L}'_{Tri}=\left[d(\mathbf{x_1},\mathbf{x_2})-d(\mathbf{p_1},\mathbf{p_2})+m\right]_+, 
\end{align}
where $\mathbf{p_1}$ and
$\mathbf{p_2}$ lie on arcs $\arc{\mathbf{x_1x_2}}$ and $\arc{\mathbf{y_1y_2}}$, respectively.
The gradients with respect to the four samples are given by:
\begin{equation}
\frac{\partial \mathcal{L}'_{Tri}}{\partial \mathbf{x_1}}=2\mathbbm{1}_{\mathcal{L}'_{Tri}>0}\left[(\mathbf{x_1}-\mathbf{x_2})-(\mathbf{p_1}-\mathbf{p_2})\cdot\left(\frac{\partial\mathbf{p_1}}{\partial{\mathbf{x_1}}}\right)\right],
\end{equation}
\begin{equation}
\frac{\partial\mathcal{L}'_{Tri}}{\partial \mathbf{x_2}}=2\mathbbm{1}_{\mathcal{L}'_{Tri}>0}\left[-(\mathbf{x_1}-\mathbf{x_2})-(\mathbf{p_1}-\mathbf{p_2})\cdot\left(\frac{\partial\mathbf{p_1}}{\partial{\mathbf{x_2}}}\right)\right],
\end{equation}
\begin{equation}
\frac{\partial\mathcal{L}'_{Tri}}{\partial \mathbf{y_1}}=2\mathbbm{1}_{\mathcal{L}'_{Tri}>0}\left[(\mathbf{p_1}-\mathbf{p_2})\cdot\left(\frac{\partial\mathbf{p_2}}{\partial{\mathbf{y_1}}}\right)\right],
\end{equation}
\begin{equation}
\frac{\partial\mathcal{L}'_{Tri}}{\partial \mathbf{y_2}}=2\mathbbm{1}_{\mathcal{L}'_{Tri}>0}\left[(\mathbf{p_1}-\mathbf{p_2})\cdot\left(\frac{\partial\mathbf{p_2}}{\partial{\mathbf{y_2}}}\right)\right].
\end{equation}

Fig. \ref{cases_fig} depicts the 9 possible ways in which $\mathbf{p_1}$ and $\mathbf{p_2}$ can be selected. Figs. \ref{ex1} and \ref{ex2} show two examples where the four samples and their updated positions obtained by using the gradients for different choices of $\mathbf{p_1}$ and $\mathbf{p_2}$ are shown. In these Figs., when $\mathbf{p_1}$ and $\mathbf{p_2}$ are not one of the end points, they are the midpoints of the arcs on which they lie. 
\par The objective of metric learning is to enable samples from the same class move towards each other and away from the samples of different classes. We use this criteria to analyze the updated samples in each case of Figs. \ref{ex1} and \ref{ex2}.
\par In Fig. \ref{ex1}, $\mathbf{x_1}$ and $\mathbf{x_2}$ are not moving towards each other in cases 2, 7, 8, whereas in cases 4, 6, 8, $\mathbf{y_1}$ and $\mathbf{y_2}$ are moving away from each other. Further, the distance between $\mathbf{y_1}$ and $\mathbf{y_2}$ remains the same in cases 0-2. Among cases 3 and 5, the distance between updated samples $\mathbf{x_1}$ and $\mathbf{x_2}$ is closer in case 5, and hence the optimal set of negatives ($\mathbf{p_1}$, $\mathbf{p_2}$) is ($\mathbf{x_1}$, $\mathbf{y_1}$). This can also be seen visually, as $\mathbf{x_1}$ and $\mathbf{y_1}$ are the closest points between the two curves.
\par In Fig. \ref{ex2}, $\mathbf{x_1}$ and $\mathbf{x_2}$ are not moving towards each other in cases 2, 7, 8, whereas in cases 3-8, $\mathbf{y_1}$ and $\mathbf{y_2}$ are moving away from each other. Among cases 0 and 1, the distance between updated samples $\mathbf{x_1}$ and $\mathbf{x_2}$ is closer in case 1, and hence the optimal set of negatives ($\mathbf{p_1}$, $\mathbf{p_2}$) is ($\mathbf{x_1}$, $\frac{\mathbf{y_1}+\mathbf{y_2}}{2}$).  This can also be seen visually, as $\mathbf{x_1}$ and $\frac{\mathbf{y_1}+\mathbf{y_2}}{2}$ are the closest points between the two curves.

\vspace{-4mm}
\section{t-SNE Visualization of the Embedding Space}
\label{tsne}
\vspace{-1mm}
We visualize the embedding space using the Barnes-Hut t-Distributed stochastic neighbor embedding (t-SNE) technique \cite{tsne}. Fig. \ref{tsne_plots} shows the t-SNE plots when the combination of LoOp and triplet loss is used for training the Cars196 dataset. It can be seen that with increasing epochs, the clusters become more distinct. Further, as training progresses, the synthetic samples (\textcolor{red}{red}) lie within the area occupied by the original samples (\textcolor{blue}{blue}).
\par It is important to note that although the epoch numbers in Fig. \ref{tsne_plots} seem large, the iterations per epoch, given by $\frac{\text{Number of classes}\times \text{Samples per class}}{\text{Batch size}}$, are small in number.
\section{Effect of Network Architecture}
\label{more_res}
In order to observe the effect of the network architecture on the proposed approach, we deploy two architectures, namely Inception-BN \cite{bn} and ResNet-50 \cite{resnet} as the feature extractors, pre-trained on the ImageNet ILSVRC dataset. In both the cases, the embeddings are l$_2$-normalized and the batch normalization layers are frozen. The learning rate is set as $10^{-5}$, and a weight decay multiplier of $4\times10^{-4}$ is used. The rest of the parameters are kept the same as mentioned in the paper. We report the NMI and Recall@K values to measure the performance. 
\begin{table}[h!]
     \begin{subtable}{\columnwidth}\centering
    {\resizebox{0.855\columnwidth}{!}{\begin{tabular}{lcccc}
    \toprule
    Method & NMI & R@1 & R@2 & R@4 \\
    \midrule
      Triplet Semi-hard \cite{triplet1} & 55.4 & 42.6 & 55.0 & 66.4 \\
      StructClustering \cite{structcluster} & 59.2 & 48.2 & 61.4 & 71.8 \\
      Proxy NCA \cite{proxynca} & 59.5 & 49.2 & 61.9 & 67.9 \\
      Binomial Deviance \cite{hist} & - & 50.3 & 61.9 & 72.6 \\
      N-pair \cite{npair} & 60.4 & 51.0 & 63.3 & 74.3 \\
      DVML \cite{dvml} & 61.4 & 52.7 & 65.1 & 75.5 \\
      Histogram \cite{hist} & - & 52.8 & 64.4 & 74.7 \\
      ECAML \cite{ecaml} & 60.1 & 53.4 & 64.7 & 75.1 \\
      Angular \cite{angular} & 61.0 & 53.6 & 65.0 & 75.3 \\
      HDC \cite{min4} & - & 53.6 & 65.7 & 77.0 \\
      EE \cite{embex} & 59.9 & 55.0 & 67.3 & 77.6 \\
      HTL \cite{htl} & - & 57.1 & 68.8 & 78.7\\
      BIER \cite{abier} & - & 57.5 & 68.7 & 78.3\\
      HTG \cite{htg} & - & 59.5 & 71.8 & 81.3\\
      ABE \cite{abe} & - & 60.6 & 71.5 & 79.8 \\
      \midrule
      LoOp-IBN (Ours) & \textbf{66.0} & 60.4 & 72.1 & 81.4\\
      LoOp-R50 (Ours) & 64.4 & \textbf{61.1} & \textbf{72.5} & \textbf{81.7}\\
    \bottomrule  
    \end{tabular}}}
    \end{subtable}
    \caption{Comparison of clustering and retrieval performance with SOTA methods for CUB-200-2011 dataset. \textbf{Bold} numbers indicate the best values. 
    - indicates not reported. IBN: Inception-BN, R50: ResNet-50.}
    \label{netcomptable}
\end{table}

\par Table \ref{netcomptable} presents the results of our approach and comparisons with state-of-the-art (SOTA) methods, like deep variational metric learning (DVML) \cite{dvml}, energy confused adversarial metric learning (ECAML) \cite{ecaml}, hierarchical triplet loss (HTL) \cite{htl}, boosting independent embeddings robustly (BIER) \cite{abier}, hard triplet generation (HTG) \cite{htg}, attention-based ensemble (ALE) \cite{abe}, as well as metric learning-based loss functions, like triplet with semi-hard mining \cite{triplet1}, proxy NCA \cite{proxynca}, N-pair \cite{npair}, histogram \cite{hist}, angular \cite{angular}. To show that improvements carry over to R-50, we run EE \cite{embex} with R-50 and compare the increase in (NMI, R@1, R@2, R@4) for LoOp vs. EE with triplet loss: (4.2, 6.0, 6.1 6.3) for GoogLeNet, (4.5, 6.1, 5.2, 4.1) for R-50. It can be seen that LoOp outperforms all the other methods for both clustering and retrieval tasks.

\section{Results for Train-Validate-Test Split}
\label{tvt}
Table \ref{tvttable} shows the comparison of results for train-validate-test \cite{mlrc} and train-test splits. These results are obtained using the same settings in Section \ref{more_res}, using ResNet-50 architecture and CUB-200-2011 dataset. 
\begin{table}[h!]
     \begin{subtable}{\columnwidth}\centering
    {\resizebox{0.865\columnwidth}{!}{\begin{tabular}{lcccc}
    \toprule
    Split & NMI & R@1 & R@2 & R@4 \\
    \midrule
      Train-Validate-Test & 59.8 & 56.4 & 68.6 & 78.9 \\
      Train-Test & 64.4 & 61.1 & 72.5 & 81.7 \\
      \bottomrule  
    \end{tabular}}}
    \end{subtable}
    \caption{Comparison of results for train-validate-test and train-test splits using CUB-200-2011 dataset.}
    \label{tvttable}
\end{table}

\end{document}